\documentclass{article}

\usepackage[nonatbib,preprint]{neurips_2023}

\usepackage[utf8]{inputenc} 
\usepackage[T1]{fontenc}    
\usepackage{url}            
\usepackage{booktabs}       
\usepackage{amsfonts}       
\usepackage{nicefrac}       
\usepackage{microtype}      
\usepackage{xcolor}         
\usepackage{graphicx}
\usepackage{float}
\usepackage{enumitem}
\usepackage{subcaption}
\usepackage[export]{adjustbox}
\usepackage{amsmath}

\usepackage{wrapfig}

\usepackage{ntheorem}
\newtheorem{hyp}{Hypothesis}

\newrobustcmd{\B}{\bfseries}

\title{Hardwiring ViT Patch Selectivity into CNNs using Patch Mixing}

\author{%
  Ariel Lee \\
  Boston University\\
  \texttt{ariellee@bu.edu} \\
  \And
  Sarah Adel Bargal \\
  Georgetown University\\
  \texttt{sarah.bargal@georgetown.edu}  \\
  \And
  Janavi Kasera \\
  Boston University\\
  \texttt{jkasera@bu.edu}  \\
  \And
  Stan Sclaroff \\
  Boston University\\
  \texttt{sclaroff@bu.edu} \\
  \And
  Kate Saenko \\
  FAIR\\
  Boston University\\
  \texttt{saenko@bu.edu}  \\
  \And
  Nataniel Ruiz \\
  Boston University\\
  \texttt{nruiz9@bu.edu} \\
}

\begin{document}

\maketitle

\begin{abstract}
Vision transformers (ViTs) have significantly changed the computer vision landscape and have periodically exhibited superior performance in vision tasks compared to convolutional neural networks (CNNs). Although the jury is still out on which model type is superior, each has unique inductive biases that shape their learning and generalization performance. For example, ViTs have interesting properties with respect to early layer non-local feature dependence, as well as self-attention mechanisms which enhance learning flexibility, enabling them to ignore out-of-context image information more effectively. We hypothesize that this power to ignore out-of-context information (which we name \textit{patch selectivity}), while integrating in-context information in a non-local manner in early layers, allows ViTs to more easily handle occlusion. In this study, our aim is to see whether we can have CNNs \textit{simulate} this ability of patch selectivity by effectively hardwiring this inductive bias using Patch Mixing data augmentation, which consists of inserting patches from another image onto a training image and interpolating labels between the two image classes. Specifically, we use Patch Mixing to train state-of-the-art ViTs and CNNs, assessing its impact on their ability to ignore out-of-context patches and handle natural occlusions. We find that ViTs do not improve nor degrade when trained using Patch Mixing, but CNNs acquire new capabilities to ignore out-of-context information and improve on occlusion benchmarks, leaving us to conclude that this training method is a way of simulating in CNNs the abilities that ViTs already possess. We will release our Patch Mixing implementation and proposed datasets for public use. Project page: \url{https://arielnlee.github.io/PatchMixing/}
\end{abstract}

\section{Introduction}

Convolutional neural networks (CNNs) and Vision Transformers (ViTs) are two dominant deep learning models for computer vision tasks. Although CNNs have established themselves as the go-to approach for many years,  the introduction of ViTs has significantly changed the landscape and they have consistently achieved comparable or superior performance compared to CNNs for key computer vision tasks such as object recognition, object detection, semantic segmentation, and many others.

In recent years, a relatively robust literature has developed comparing CNNs and Vision Transformers in terms of overall performance on standard benchmarks, robustness to OOD inputs, robustness to adversarial attacks, and other evaluations~\cite{Mahmood_2021_ICCV,wang2022can,borji2022cnns,matsoukas2021time,NEURIPS2021_e19347e1,ruiz2022finding,convnext,ranzato2021vision}, as well as analysis work that compares the way both architecture types understand images and how they ultimately arrive at their predictions~\cite{ruiz2022finding,ranzato2021vision,intriguing_transformers,pinto2022impartial}. 

We note that one important research topic remains under-explored: how these architectures handle occlusion. There exists work that compare both architectures using simple simulations of occlusion such as patch dropping~\cite{intriguing_transformers}, or occlusion in a simulated environment~\cite{ruiz2022finding}. Additionally, in work by Pinto et al.~\cite{pinto2022impartial}, they found no clear winner between modern CNNs and ViTs for different robustness tests. In this work, we dive deeply into this specific area and present four main contributions:

\begin{itemize}
    \item We find a previously undiscovered \textbf{incontrovertible difference} in performance between modern ViTs and CNNs. ViTs are naturally more robust when \textit{out-of-context information} is added to an image compared to CNNs. We call this ability to ignore out-of-context patches: \textit{patch selectivity}.
    \item We revisit \textbf{Patch Mixing}, a data augmentation method where patches from other images are introduced into training images and ground-truth labels are interpolated. We show that by training CNNs using Patch Mixing, we \textit{simulate} the natural ability of ViTs to \textbf{ignore out-of-context information}.
    \item We show that models with better patch selectivity tend to be \textbf{more robust to natural occlusion}. Specifically, we introduce two new challenging datasets to evaluate performance of image classifiers under occlusion: the Superimposed Masked Dataset (SMD) and the Realistic Occlusion Dataset (ROD). Moreover, our CNN models trained using Patch Mixing become more robust to occlusion in these, and other datasets.
    \item We propose a new explainability method, \textbf{c-RISE} - a contrastive version of the RISE~\cite{Petsiuk2018rise} explainability method that allows for agnostic analysis of input sensibility under occlusion for both CNNs and Transformers. Using c-RISE we are able to measure patch selectivity and show that augmentation using Patch Mixing improves CNN patch selectivity.
\end{itemize}

\begin{figure}
    \centering
    \includegraphics[width=1.0\textwidth]{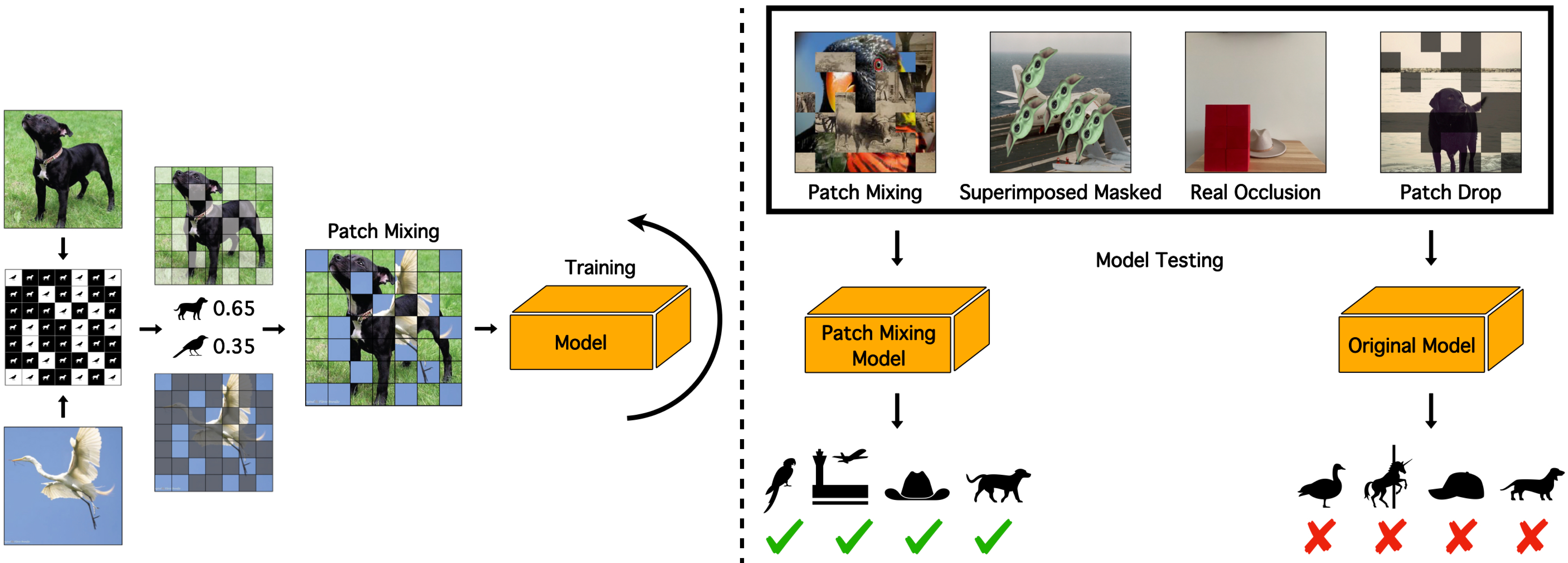}
    \caption{Patch Mixing augmentation with label smoothing improves the ability of CNNs to handle a multitude of alterations and occlusions, bridging the gap with ViTs.}
    \label{fig:illustration}
\end{figure}

\section{Deep Dive Into Patch Selectivity}

\paragraph{Modern CNN and ViT Inductive Biases}
Convolutional neural networks (CNNs) are traditionally composed of a series of trainable convolutional layers. Modern CNN architectures such as ConvNeXt~\cite{convnext} differ in many respects, yet still follow a purely convolutional approach. A particularly important change is the use of a patchify stem - this change can both increase the overall receptive field in early layers in modern convnets as opposed to traditional convnets, as well as decrease strong local dependencies that are created in the early layers of the network, since the patches are non-overlapping. Nevertheless, this, and other changes, do not completely change the inductive bias of the architecture: the network remains a purely convolutional network that uses square conv filters, has a propensity to more strongly weight proximal evidence, and has relatively small effective receptive fields in early layers.

The Vision Transformer (ViT)~\cite{dosovitskiy2020image} is a neural network architecture for image recognition that uses self-attention based Transformer layers. An image is first divided into non-overlapping patches, that are then transformed into embeddings. These embeddings are used as inputs for the Transformer layers. ViTs possess distinct properties and inductive biases when compared to CNNs, some of which are particularly important to highlight.

\paragraph{ViT Early Layer Long-Range Dependence}
In CNNs the receptive field at a specific layer is fully determined by the size of the convolution kernel and the stride of that layer, as well as the layers that precede the layer of interest. For this reason, given limits on the convolutional kernel size and the stride of the kernel, the receptive field for early CNN layers does not encompass the full image input. In contrast, early layers of ViTs have a large receptive field because they use self-attention, which allows them to attend to any part of the input image beginning at the first layer of the architecture. As a result, ViTs can learn relationships between pixels that are far apart in the input image~\cite{ranzato2021vision}, while CNNs are limited to learning relationships between proximal pixels. In this way, ViTs have the property of early-layer long-range dependency that is not possible to structurally mimic in CNNs, even with modernized CNN architectures that include patchify stems.
In this work we pose the following:
\begin{hyp}
\label{hyp:1}
Hierarchical attention in ViT-style networks allows them to more easily discount signal from out-of-context information in an image when compared to CNNs, which, due to their structure and inherent inductive biases, have a harder time discounting signal from out-of-context patches.
\end{hyp}

Specifically, in this work we evaluate this hypothesis using empirical means. This hypothesis has been discussed in the prominent work of Naseer et al.~\cite{intriguing_transformers} that compares ViT and CNN performance when faced with occlusion. They study occlusion by simulating it using either random or saliency-guided patch dropping in images. In particular, the main conclusion is that ViTs were vastly better at dealing with out-of-context patches. Nevertheless, this study focused on older convnet architectures such as ResNet50, DenseNet121 and VGG19. Modern convnets such as ConvNeXt, proposed in the influential work of Liu et al.~\cite{convnext}, possess very different architectures while remaining fully-convolutional. There is a relative scarcity of study of these new architectures with respect to occlusion, although recent work~\cite{ruiz2022finding} proposes to study occlusion for Swin Transformers and ConvNeXt CNNs. Interestingly, they find that new innovations in architecture and training regime makes these new convnets much stronger than older convnets such as ResNet50 at ignoring dropped patches, yet still lagging behind ViTs at higher levels of information loss. 

One important issue to raise, is that patch drop is a poor approximation of real world occlusion, where occluders are usually other objects that have their own shape and texture, which adds another dimension to the problem. The question then remains: 
\textit{Are ViTs truly better at handling occlusion and discounting signal from out-of-context patches than CNNs?}

\begin{figure}[ht!]
  \centering
  \includegraphics[width=0.28\textwidth]{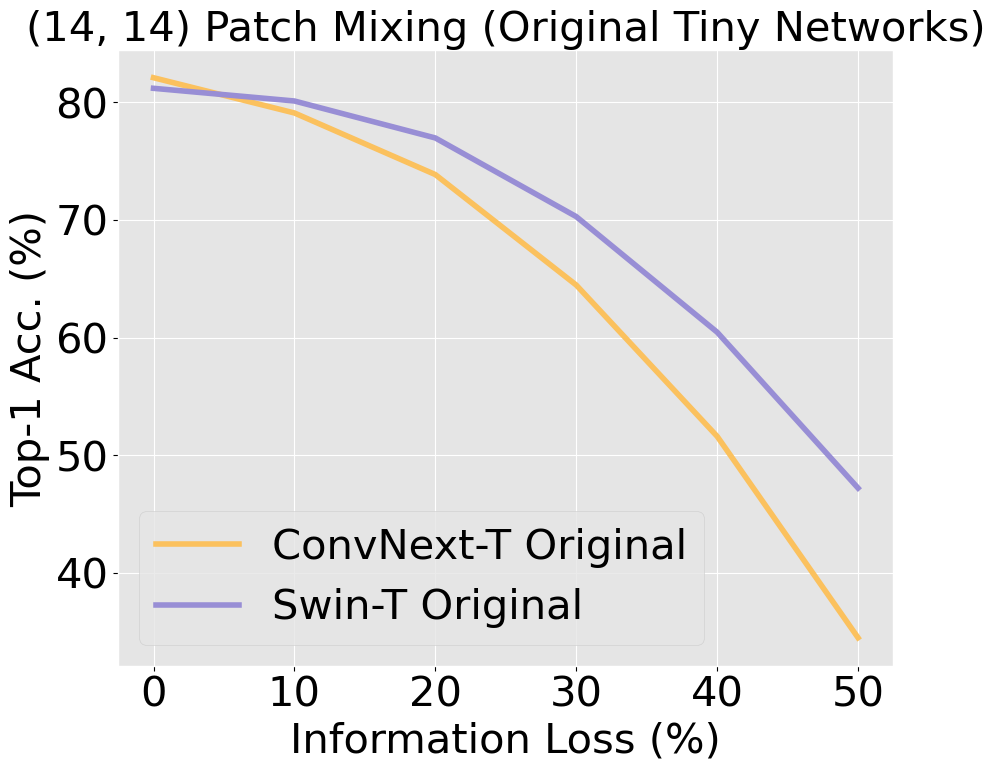}
  \includegraphics[width=0.28\textwidth]{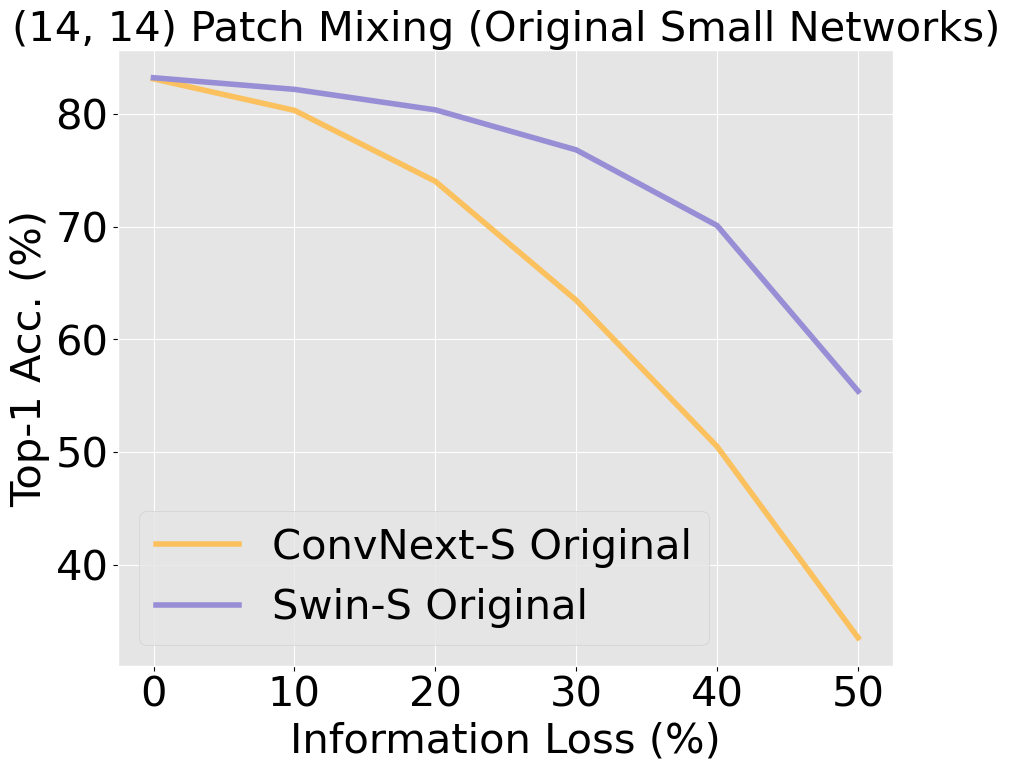}
  \caption{ConvNeXt performance severely decreases as more out-of-context patches are inserted into test images, with Swin proving to be more resilient to this type of occlusion.}
  \label{fig:patch_mixing_method_fig}
\end{figure}

We find that \textbf{the answer is a resounding yes}. Specifically, when comparing ViTs and modern convnets that have identical parameter count, FLOPs and very close ImageNet validation performance, ViTs degrade much less when out-of-context patches are introduced into an image. In Figure~\ref{fig:patch_mixing_method_fig}, we show the accuracy of comparable ConvNeXt and Swin models when out-of-context patches are introduced into test images. We see a much larger decrease in accuracy in ConvNeXt compared to Swin, with a widening gap as information loss increases. This finding is particularly interesting in the context of recent work by Pinto et al.~\cite{pinto2022impartial}, which finds no clear winner in a contest between ConvNeXt and Swin models of different sizes for different robustness tests such as simplicity bias, background bias, texture bias, OOD detection and other tasks. To the best of our knowledge we are the first to find an incontrovertible difference between these two classes of models that stands out. 

This experiment is a rough approximation of natural occlusions, where objects or surfaces occlude the main object in an image. We do, however, hypothesize that networks that can more easily discount signal from out-of-context patches will tend to perform better under naturalistic occlusion:
\begin{hyp}
\label{hyp:2}
A model with better patch selectivity will tend to perform better under naturalistic occlusion.
\end{hyp}

In order to test this, we first evaluate the patch selectivity of our trained models, and then extensively test them on four different benchmarks, including two datasets that we propose as contributions: the Superimposed Masked Dataset (SMD) and the Realistic Occlusion Dataset (ROD) which will be described further below. We find that there is indeed a positive correlation between patch selectivity and performance under occlusion, and supply the details in the Experiments section.

Finally, we pose the following final hypothesis:
\begin{hyp}
\label{hyp:3}
A model that is explicitly trained to deal with out-of-context patches using data augmentation will tend to improve at ignoring out-of-context information at test-time.
\end{hyp}
In our experiments we evaluate this hypothesis and show that using Patch Mixing at training time improves CNN patch selectivity, but, surprisingly, does not improve ViT patch selectivity. We believe this is due to the fact that patch selectivity is already a natural capability of ViTs, whereas CNNs have lesser patch selectivity and attention \textit{bleeds out} from in-context patches to neighbouring out-of-context patches. By combining verified Hypotheses \ref{hyp:2} and \ref{hyp:3}, we can conclude that CNNs trained using Patch Mixing are more robust to natural occlusions in the real world. We indeed confirm this experimentally.

\subsection{Augmentation by Patch Mixing}
Previous work has introduced the notion of inserting parts of different images into training images in different manners. CutMix~\cite{yun2019cutmix} proposes to cut and paste one contiguous rectangle from another image into a training image, and mix the ground truth labels proportionally to the area of each image. Cascante-Bonilla et al.~\cite{cascante2021evolving} propose random and evolutionary search-guided replacement of training image square patches with patches from another training image, also mixing ground truth labels in proportional fashion. \cite{xu2022patchmix} proposes replacing rectangular patches in an image, with patches from many other training images, in order to augment small datasets in few-shot learning.

Our proposed augmentation is named Patch Mixing. Let $x \in \mathbb{R}^{H \times W \times C}$ and $y$ denote the image and its label respectively. We seek to generate an image/label pair $(\tilde{x}, \tilde{y})_i$ by mixing patches from images $x_A$ and $x_B$ while appropriately mixing labels $y_A$ and $y_B$. For this we generate a mask composed of patches $M \in  \{0, 1\}^{N \times P^2 \times C}$, where $(H, W)$ is the resolution of the original image, $C$ is the number of channels, $(P, P)$ is the resolution of each image patch, and $N = \frac{HW}{P^2}$ is the resulting number of patches. We initialize the elements of this mask to $0$. We then select $N_1$ patches from this mask, following uniform random sampling and set the elements of those patches to $1$. These are the patches that will be replaced in image $x_A$. We select $N_1$ based on a proportion hyperparameter $r = N_1 / N$ which represents the proportion of patches that are replaced. Finally, we generate $\tilde{x}$:
\begin{equation}
    \tilde{x} = (1 - M) \odot x_A + M \odot x_B.
\end{equation}
Labels $y_A$ and $y_B$ are mixed to generate label $\tilde{y}$, using the proportion $r$. The resulting vector is smoothed using label smoothing~\cite{szegedy2016rethinking}.

Our proposed Patch Mixing most resembles one method mentioned in \cite{cascante2021evolving}, with some important differences in both application scenario and implementation. For the application scenario, their work does not study the effects of Patch Mixing on Transformers, doing so only on CNNs. Moreover, they solely study ResNet and MobileNet architectures, and the method was not applied to modern convnets given  the concurrency of \cite{convnext} and their work. Finally, most evaluations in their work are based on the CIFAR-10 dataset~\cite{krizhevsky2014cifar}, while we evaluate improved networks on four datasets that present different types of occlusion simulations and real-world occlusions.

Our Patch Mixing implementation has important differences with \cite{cascante2021evolving}. First, we find that in order to recover the strong performance exhibited by modern CNNs on ImageNet it is imperative to \textit{disable} random erasing when using patch mixing. When both are used simultaneously, information loss is too high, resulting in lower overall performance. Next, our version uses label smoothing~\cite{szegedy2016rethinking} which increases performance. We also find that using a more granular grid for patch replacement improves results for modern CNNs - thus we use a 7x7 grid instead of a 4x4 grid. Their work focuses on a guided version of mixing patches using evolutionary search. We find that random patch mixing is less computationally expensive and suffices to evaluate the hypotheses of this work.

\subsection{Contrastive RISE (c-RISE) and Patch Selectivity}

Petsiuk et al.~\cite{Petsiuk2018rise} proposed Randomized Input Sampling for Explanation of Black-box Models (RISE), a method that generates an image heatmap that highlights the importance of pixel evidence in that image for a specific prediction $y_\text{pred}$. This method is a perfect fit for our problem since it is an empirical method that is model agnostic and can be applied to both modern CNNs and ViTs. Specifically, it uses iterative random masking of an image using Monte Carlo sampling, and evaluates the predictions of the model on the masked images to generate an importance map. Unfortunately, RISE is not a contrastive method that generates evidence maps for a specific class, and only that class. This is a direly needed property for us, since occluders can be in the label space of the model, which can cause them to be highlighted as non-specific evidence using traditional RISE. We propose a grey-box modification of RISE called contrastive RISE (c-RISE), where the Monte Carlo equation becomes:
\begin{equation}
S_{x, f}(\lambda) \overset{\mathrm{MC}}{\approx} \frac 1{\mathbb{E}[B]\cdot N_B}\sum_{i=1}^{N_B} [f(x\odot B_i) - f^{\prime}(x\odot B_i)]\cdot B_i(\lambda).
\end{equation}
Where $B_i$ is the sample binary mask, and $f^{\prime}$ is the classifier $f$ with the weights of the last fc layer flipped (multiplied by $-1$) following the trick proposed in \cite{zhang2018top}. For more information on c-RISE please refer to the supplementary material.

Finally, we present an empirical approximation of patch selectivity using c-RISE, which corresponds to the contrastive importance of in-context areas of the image. Simply, we sum the parts of the c-RISE importance heatmap that overlap with image patches that are from the original image (and not from the occluder image):
\begin{equation}
\label{eq:patch_selectivity}
\mathcal{P}_{f}(x) = \frac{1}{N} \sum S_{x,f} \odot (1 - M).
\end{equation}
\section{Datasets}

\paragraph{Realistic Occlusion Dataset (ROD)}

The Realistic Occlusion Dataset is the product of a meticulous object collection protocol aimed at collecting and capturing 40+ distinct objects from 16 classes: \textit{banana, baseball, cowboy hat, cup, dumbbell, hammer, laptop, microwave, mouse, orange, pillow, plate, screwdriver, skillet, spatula, and vase}. Images are taken in a bright room with soft, natural light. All objects are captured on a brown wooden table against a solid colored wall. An iPhone 13 Pro ultra-wide camera with a tripod is used to capture images at an elevation of approximately 90$^\circ$ and distance of 1 meter from the object. Occluder objects are wooden blocks or square pieces of cardboard, painted red or blue. The occluder object is added between the camera and the main object and its x-axis position is varied such that it begins at the left of the frame and ends at the right. In total, 1 clean image and 12 occluded images are captured for each object. Each object is measured and the occluder step size is broken up into equal sizes. 

\begin{figure}
    \centering
\resizebox{1.0\textwidth}{!}{%
    \includegraphics[width=.14\textwidth, frame]{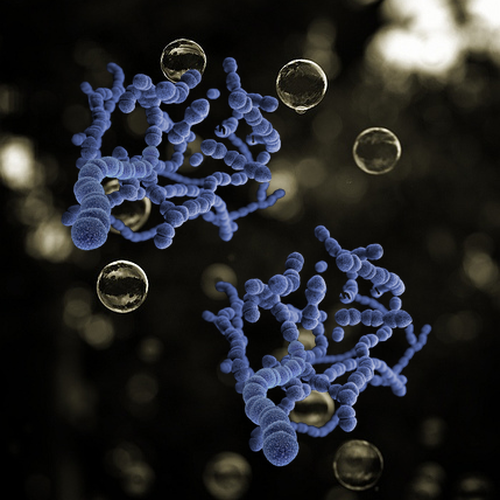}
    \includegraphics[width=.14\textwidth, frame]{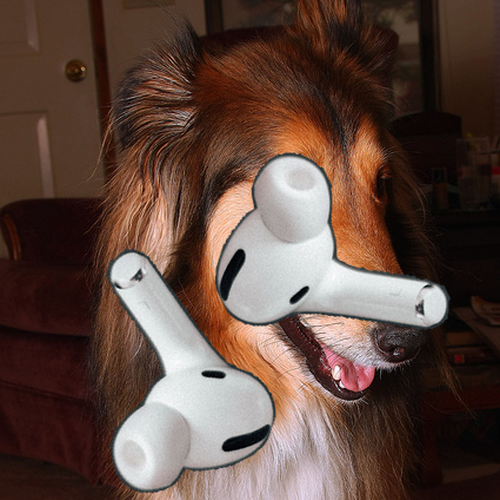}
    \includegraphics[width=.14\textwidth, frame]{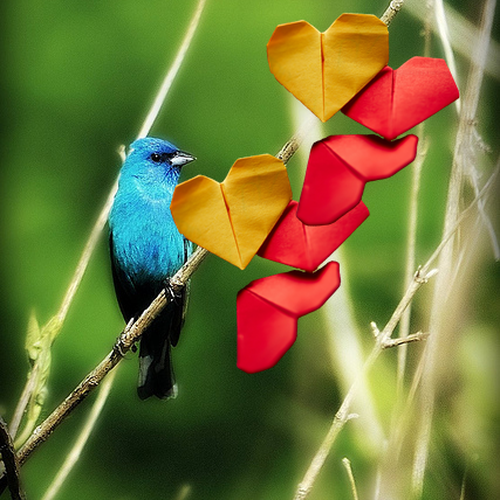}
    \ \
    \includegraphics[width=.14\textwidth, frame]{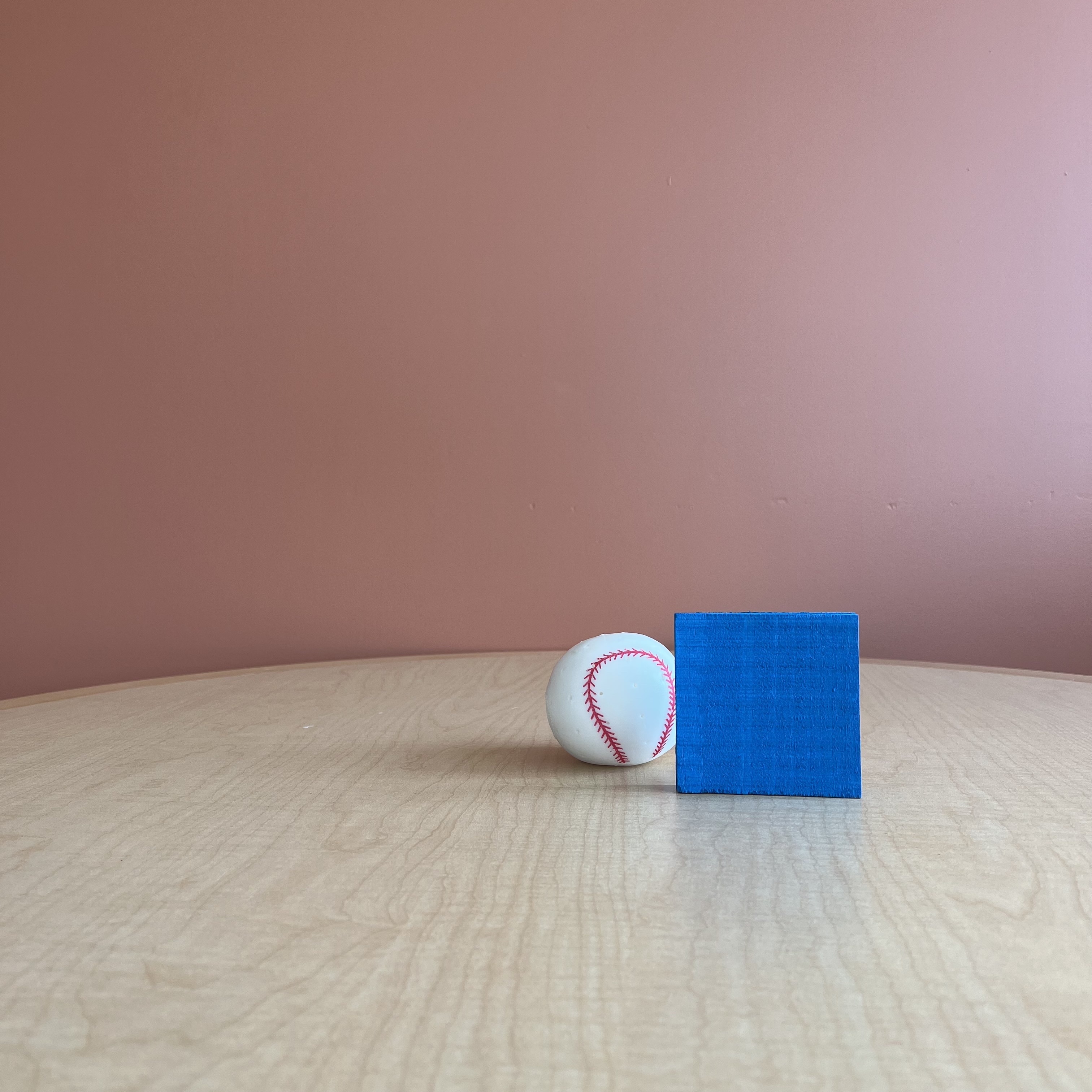}
    \includegraphics[width=.14\textwidth, frame]{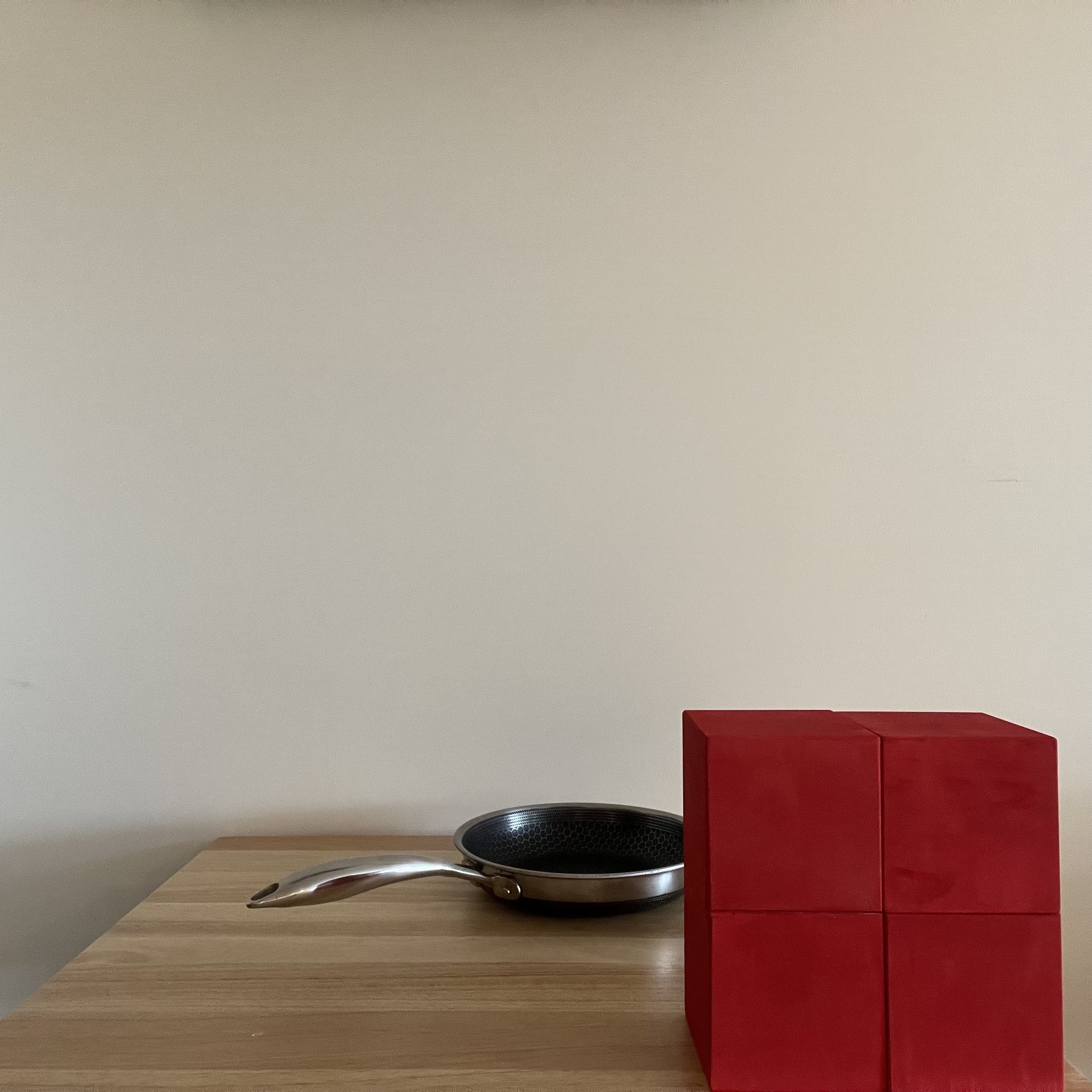}
    \includegraphics[width=.14\textwidth, frame]{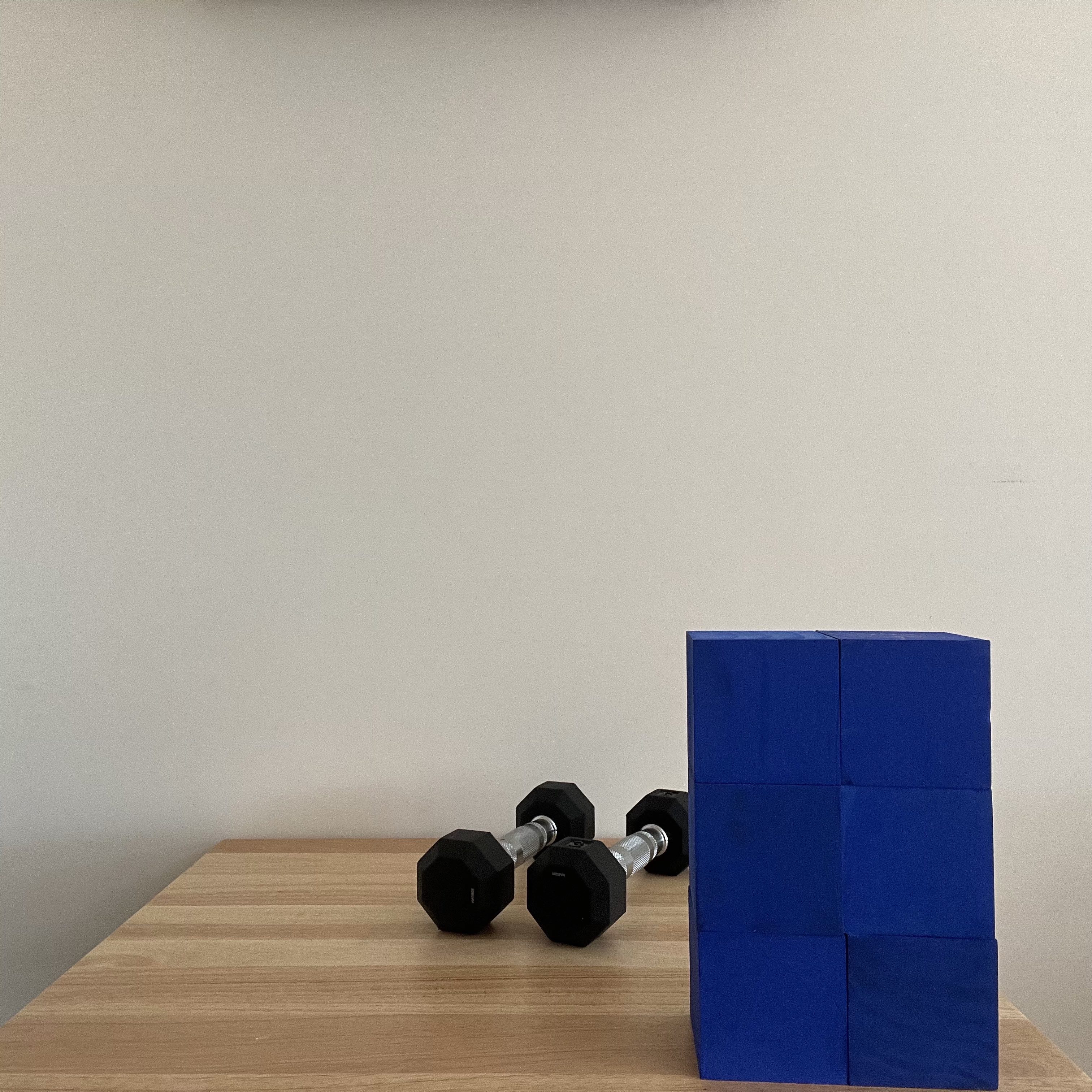}
    }
    \caption{Random examples from our proposed challenging occlusion datasets: SMD (left 3 images) and ROD (right 3 images) datasets.}
    \label{fig:occ_examples}
\end{figure}

\paragraph{Superimposed Masked Dataset (SMD)}
We generate three versions of SMD, an occluded ImageNet-1K validation set, as an additional way to evaluate the impact of occlusion on model performance. This experiment used a variety of occluder objects that are not in the ImageNet-1K label space and are unambiguous in relationship to objects that reside in the label space. Two occluder objects for each of the following classes were segmented using Meta's Segment Anything~\cite{kirillov2023segment}: \textit{airpods, virtual reality headset, drone, graduation cap, anatomical heart, origami heart, skateboard, diamonds (stones, not in a setting), Grogu (baby yoda), person, popcorn, coronavirus, bacteriophage, and bacteria}. Figure~\ref{fig:occ_examples} shows examples of images from the SMD datasets with varying levels of occlusion.

\section{Experiments}

\paragraph{Models and Training}
The Patch Mixing models are trained from scratch using the original training scripts. The only hyperparameter change made is the removal of random erasing. When augmenting, we set an equal probability of using Mixup, CutMix, or Patch Mixing. For each batch of images, the patching ratio is randomly sampled from a beta distribution. If not specified, experiments are conducted on the ImageNet validation set. Tiny networks were trained on 4 RTX8000 and Small networks on 4 A6000.

\subsection{Patch Selectivity}

\begin{figure}
    \centering
    \resizebox{0.9\textwidth}{!}{%
    \includegraphics[width=0.3\linewidth]{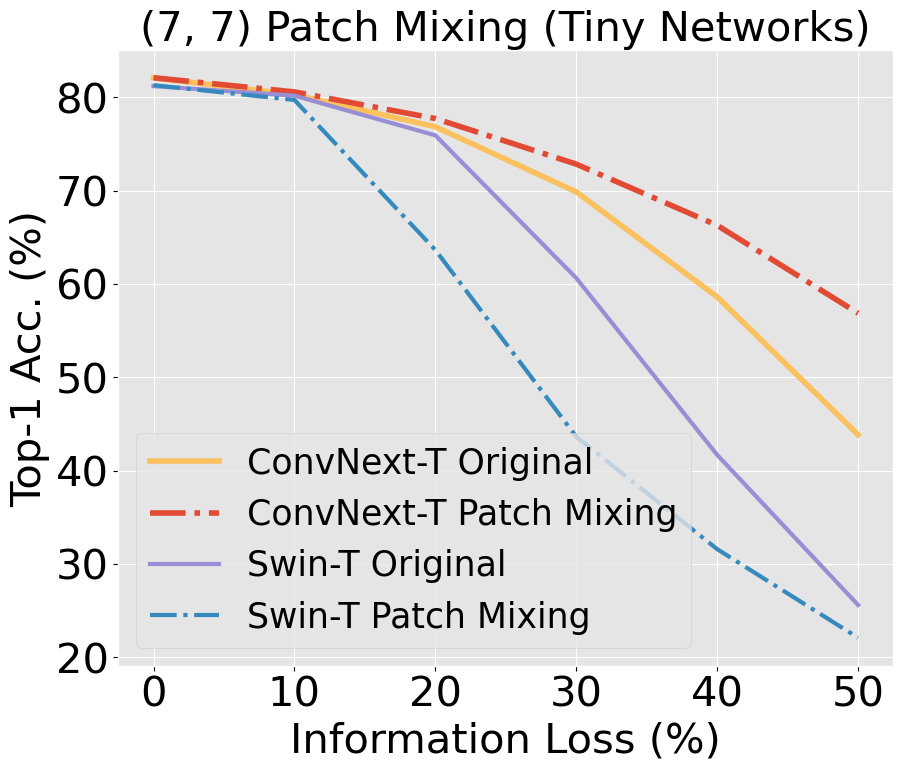}
    \includegraphics[width=0.3\linewidth]{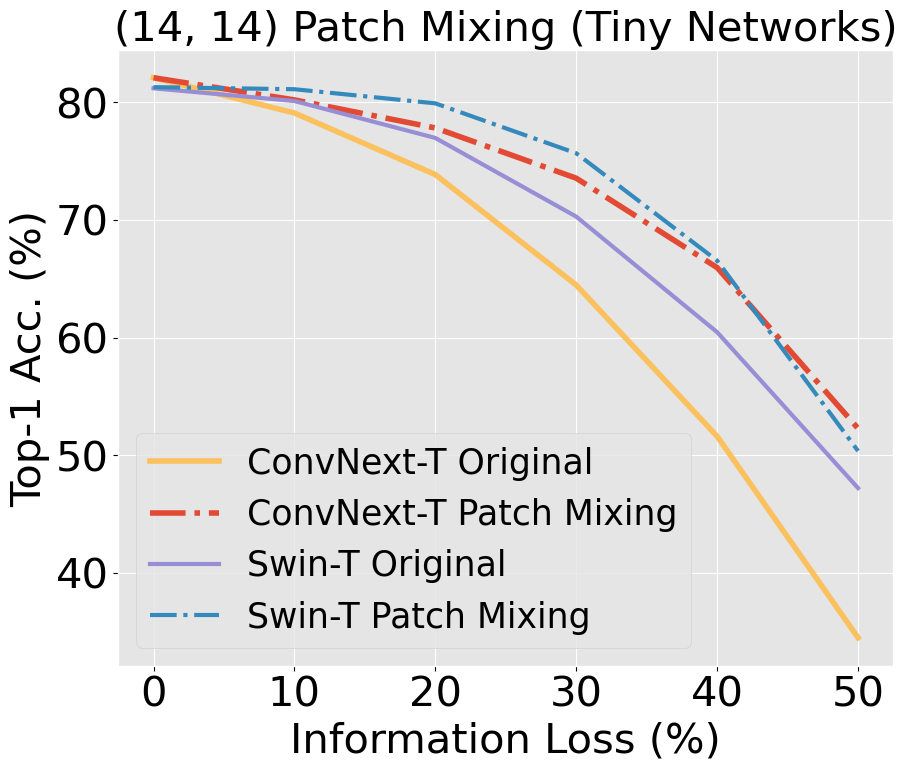}
    \includegraphics[width=0.3\linewidth]{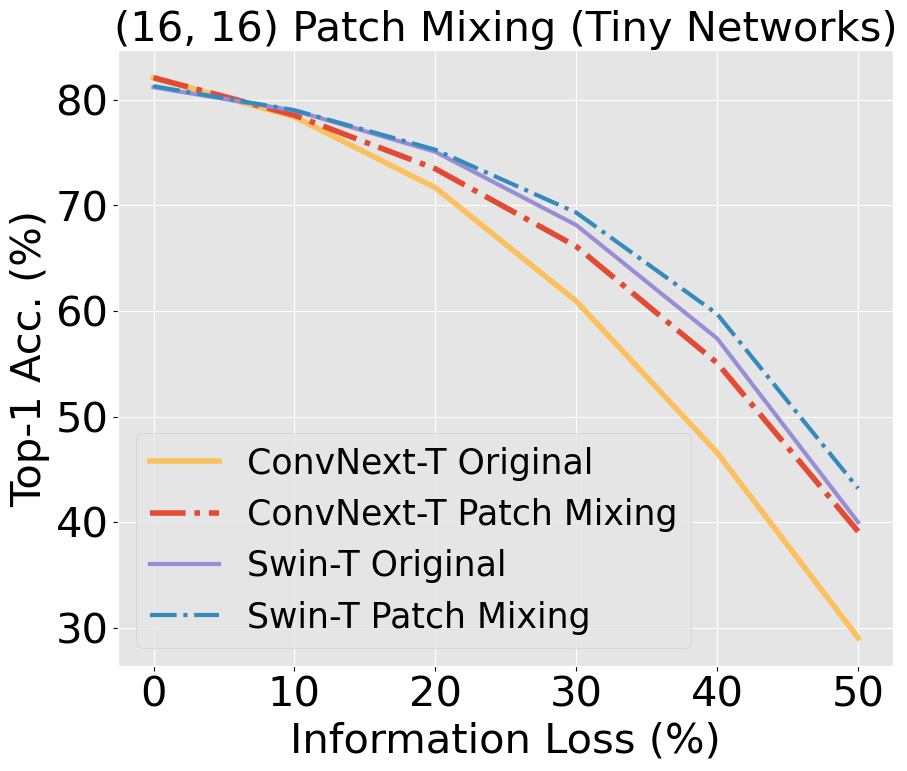}
    }
    \\
    \resizebox{0.9\textwidth}{!}{%
    \includegraphics[width=0.3\linewidth]{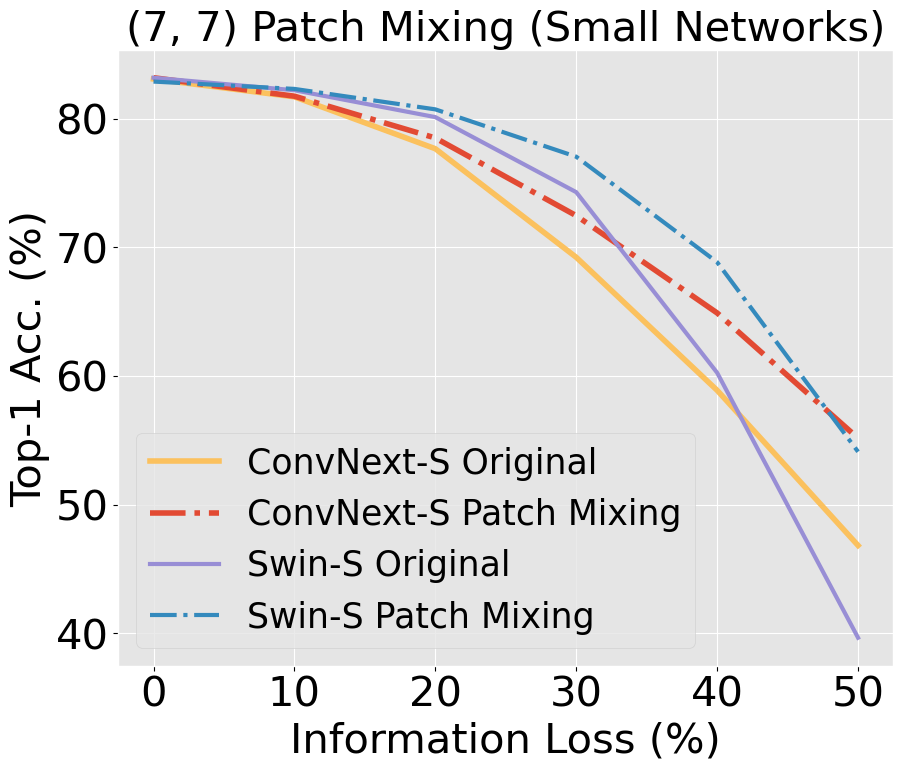}
    \includegraphics[width=0.3\linewidth]{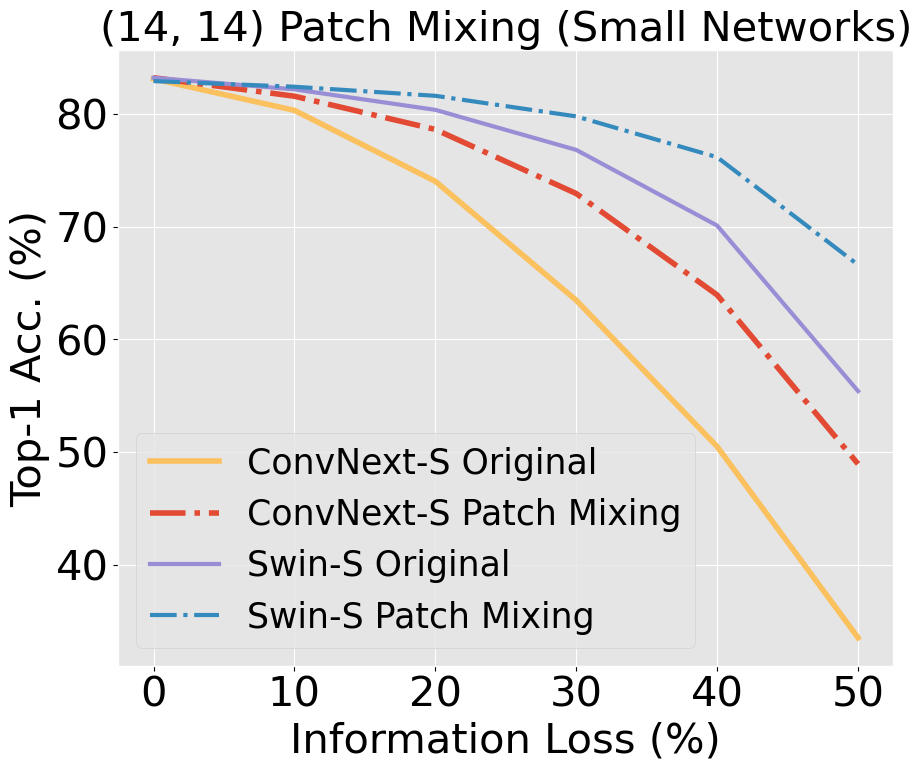}
    \includegraphics[width=0.3\linewidth]{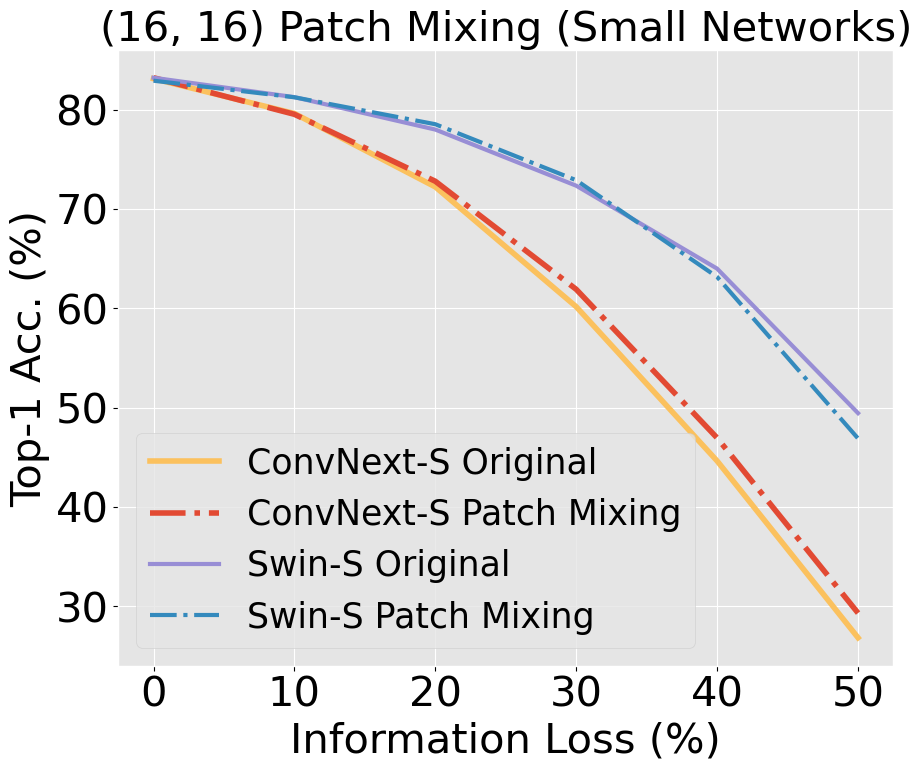}
    }
    \caption{Patch Mixing experiments on tiny and small networks on ImageNet-1K val. \textit{ViTs natively have better patch selectivity than CNNs, yet when we use Patch Mixing augmentation, CNNs have similar patch selectivity to ViTs.}}
    \label{fig:Patch_Mixing_tiny}
\end{figure}

\paragraph{ViTs have better patch selectivity than CNNs}
To test a model's ability to ignore out-of-context patches, we run patch mixing experiments on ImageNet-1K val and report the Top-1 accuracy as a function of information loss in Figures~\ref{fig:Patch_Mixing_tiny}. Note that no label smoothing is applied for attacked images and the information loss degree is deterministic. We present different experiments using different number of image patches. We observe that Original Swin models vastly outperform Original ConvNeXt models as information loss increases. Specifically, this shows that Swin can naturally ignore out-of-context patches better than ConvNeXt.

\paragraph{Using Patch Mixing augmentation, CNNs have similar patch selectivity to ViTs}
By examining Figures~\ref{fig:Patch_Mixing_tiny}, we can see that using patch mixing augmentation ConvNeXt equals the performance of original Swin with respect to patch replacement attacks, gaining the ability of patch selectivity that ViTs inherently have. To add further evidence to this, Swin networks do not improve much on average using patch mixing, which suggests that we are supplying an inductive bias that is already present in the architecture.
\subsection{Spatial structure invariance}

\begin{figure}
    \centering
    \begin{minipage}{0.35\linewidth}
        \includegraphics[width=\linewidth]{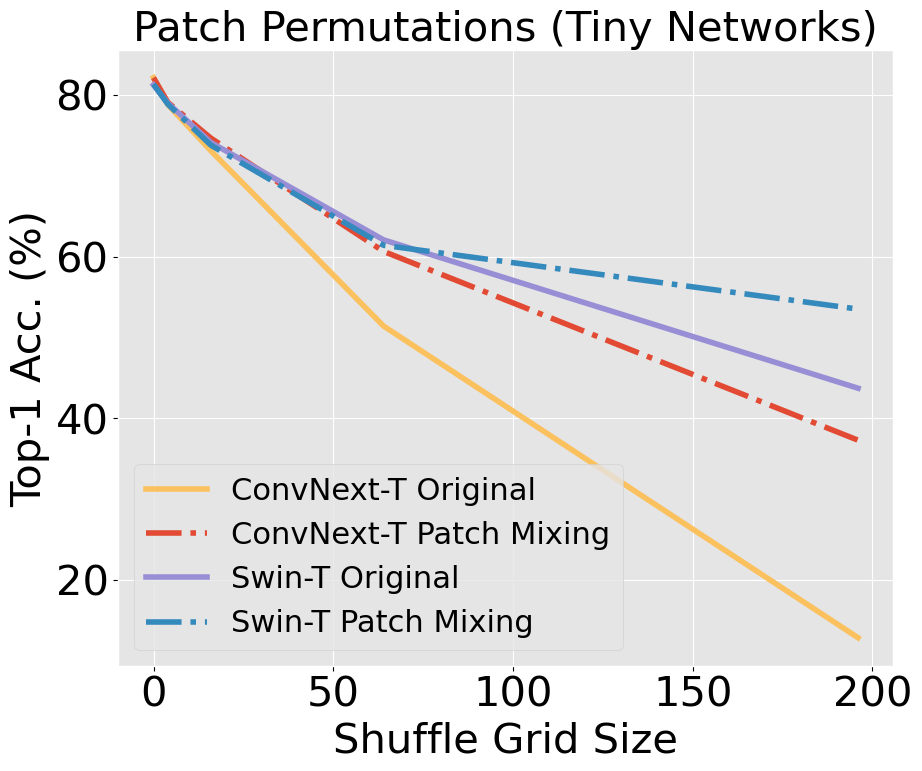}
        \subcaption{}
    \end{minipage}
    \begin{minipage}{0.35\linewidth}
        \includegraphics[width=\linewidth]{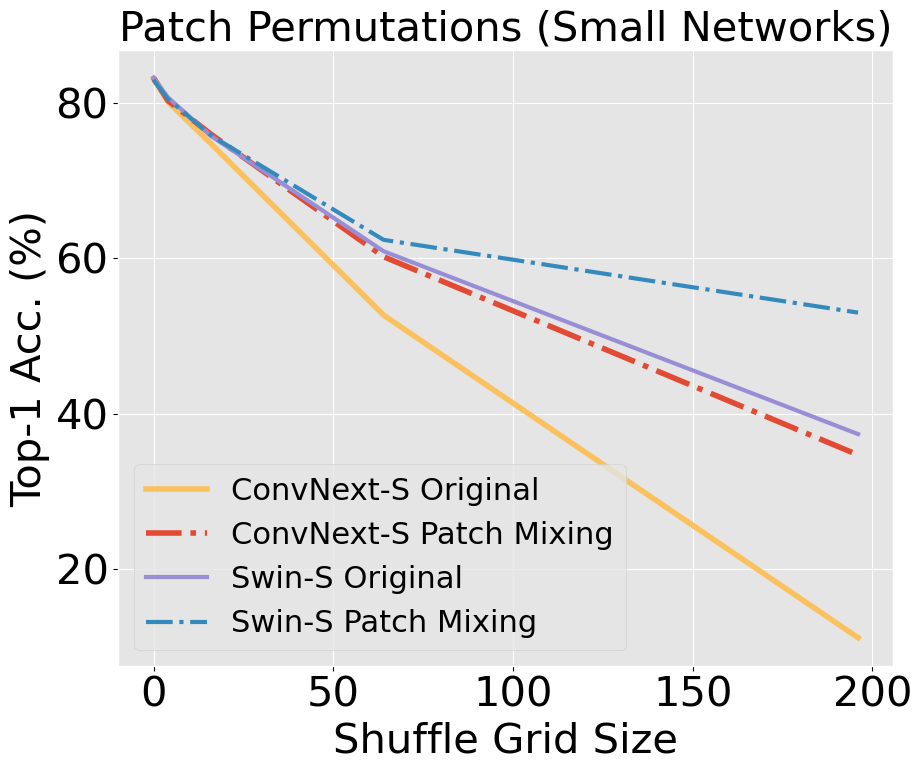}
        \subcaption{}
    \end{minipage}
    \
    \
    \begin{minipage}{0.15\linewidth}
        \includegraphics[width=\linewidth, frame]{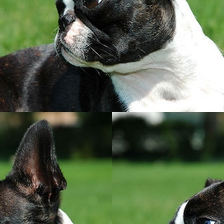}
        \subcaption{2x2 grid} 
        \vspace{3pt}
        \includegraphics[width=\linewidth, frame]{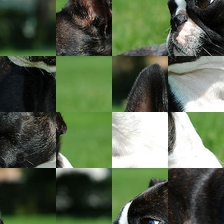}
        \subcaption{4x4 grid}
    \end{minipage}
    \caption{\textbf{Better patch selectivity means greater resistance to abnormal spatial structure:} Top-1 accuracy on IN-1k val set is plotted against shuffle grid size for the patch permutation experiments on Tiny (a) and Small (b) networks. Examples of patch permutations can be seen in (c) and (d).}
    \label{fig:perms}
\end{figure}

\paragraph{Patch Mixing bestows better spatial structure invariance to CNNs} The fundamental architecture of ViTs offers inherent, "out-of-the-box" permutation invariance. We re-implement the patch permutation experiments conducted in \cite{intriguing_transformers} and find that, surprisingly, Patch Mixing reduces modern CNNs reliance on spatial structure, resulting in context-independence and robustness to permutations on par with ViT models. In Figure~\ref{fig:perms} we see that the performance gap between original and Patch Mixing trained ConvNeXt models increases with the shuffle grid size. Conversely, the performance gap between ConvNeXt-T trained with Patch Mixing and the original Swin-T network remains small even as the shuffle grid size increases. The accuracy of ConvNeXt-S patch is nearly identical to the original Swin-S network. Interestingly, this is the only experiment where Swin trained with Patch Mixing shows a consistent improvement over its original counterpart.

\subsection{Robustness to occlusion}

\paragraph{Patch Mixing improves robustness to occlusion for CNNs but not for ViTs}
Table \ref{summary} presents a summary of the results for different network architectures tested on three datasets: ImageNet-1K val (IN) top-1, SMD top-1 (avg. over 10-30\% occlusion), NVD~\cite{ruiz2022finding} simulated occlusion validation top-5, and ROD top-5. The ConvNeXt and Swin networks are compared in their standard and Patch versions, both in Tiny (T) and Small (S) configurations. In the Tiny category, ConvNeXt-T and ConvNeXt-T Patch Mixing both achieved an IN top-1 score of 82.1\%, but the Patch Mixing version performed better in the NVD occlusion set (26.1\% vs. 25.4\%), SMD (48.9\% vs. 47.6\%), and ROD (42.6\% vs. 40.4\%). For the Swin-T versions, the Patch Mixing model showed minor improvements over the original in the IN and NVD occlusion datasets but slightly under-performed on ROD. The trend is mirrored for Small models.

Overall, the table suggests that the Patch variants of CNNs generally showed improved performance on occluded datasets compared to their original counterparts, whereas ViTs do not substantially improve.

\begin{table}
\caption{Mean accuracy results for IN, ROD, SMD, and NVD test sets (\%).} 
\label{summary}
\centering
\resizebox{0.5\textwidth}{!}{%
    \begin{tabular}{lllll}
    \toprule
    \multicolumn{1}{c}{Architecture} & IN & SMD & NVD & ROD \\ 
    \toprule
    ConvNeXt-T Original & 82.1 & 47.6 & 25.4 & 40.4 \\ 
    ConvNeXt-T Patch Mixing & 82.1 & \B 48.9 & \B 26.1 & \B 42.6 \\ 
    \midrule
    ConvNeXt-S Original & 83.1 & 49.4 & 21.9 & 48.4 \\ 
    ConvNeXt-S Patch Mixing & \B 83.2 & \B 50.1 & \B 25.8 & 48.4 \\ 
    \midrule
    \midrule
    Swin-T Original & 81.2 & 56.5 & 18.4 & \B 41.9 \\ 
    Swin-T Patch Mixing & \B 81.3 & \B 57.2 & \B 18.9 & 40.2 \\ 
    \midrule
    Swin-S Original & \B 83.2 & \B 60.4 & \B 20.5 & 44.3 \\ 
    Swin-S Patch Mixing & 82.9 & 60.2 & 18.2 & \B 48.2 \\ 
    \bottomrule
    \end{tabular}
}
\end{table}

\begin{figure}
    \centering
    \resizebox{0.8\textwidth}{!}{%
    \includegraphics[width=0.3\linewidth]{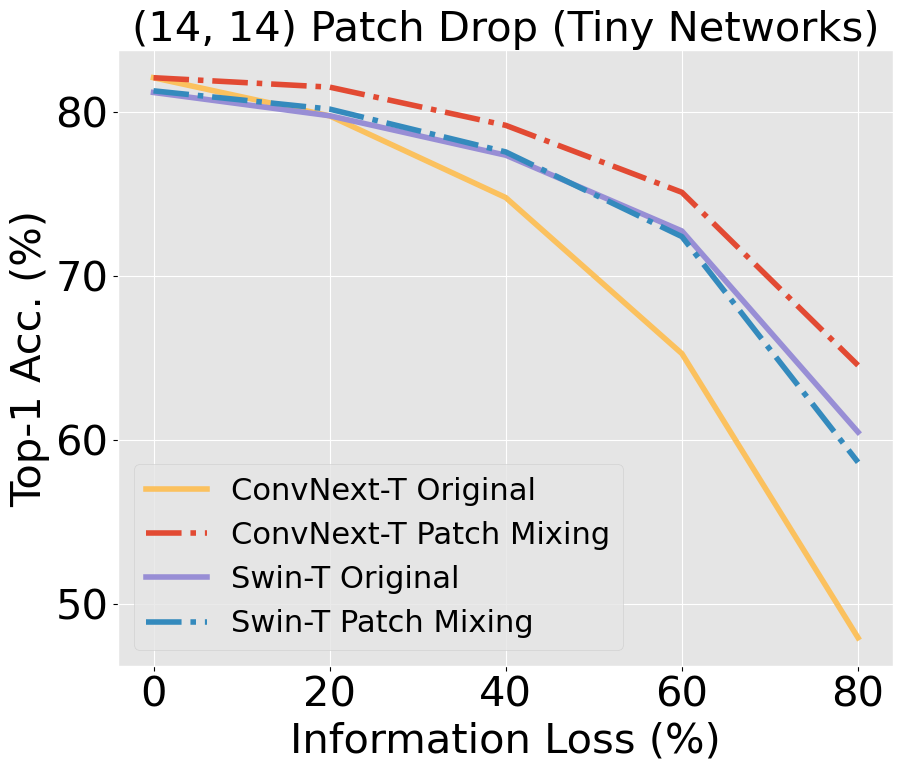}
    \includegraphics[width=0.3\linewidth]{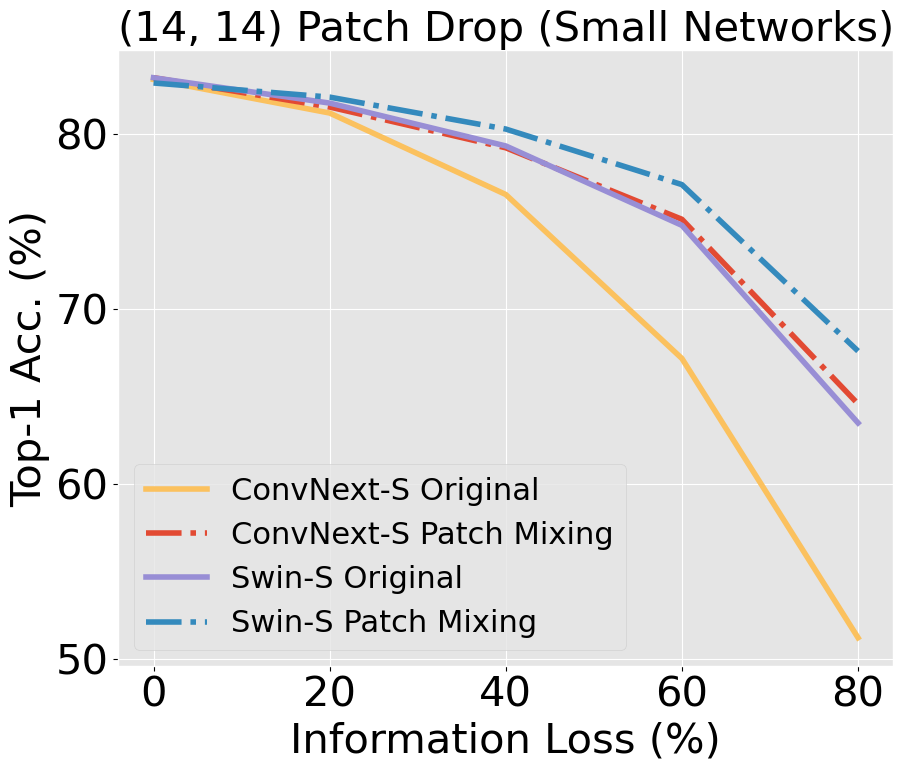}
    }
    \caption{Random patch drop: Tiny and Small networks}
    \label{fig:small_drop}
\end{figure}

\paragraph{Random Patch Drop}
Figure~\ref{fig:small_drop} illustrates that for tiny and small networks with grid size (14, 14) ConvNeXt trained with Patch Mixing outperforms its counterpart, and in some cases achieves the best result with increasing information loss. We also see that Swin performance either stays static or slightly increases, but not by the same magnitude as ConvNeXt performance.

\paragraph{c-RISE}

We obtain c-RISE maps from images that are attacked using patch mixing for both original and improved ConvNeXt and Swin models. We normalize the importance map using a Softmax function and calculate the inverse of our defined patch selectivity metric in Equation \ref{eq:patch_selectivity} by summing the importance values in out-of-context patches. To obtain granular heatmaps we increase the number of RISE masks to 14,000 and use a stride of 14.

\begin{figure}
    \centering
    \includegraphics[width=0.9\linewidth]{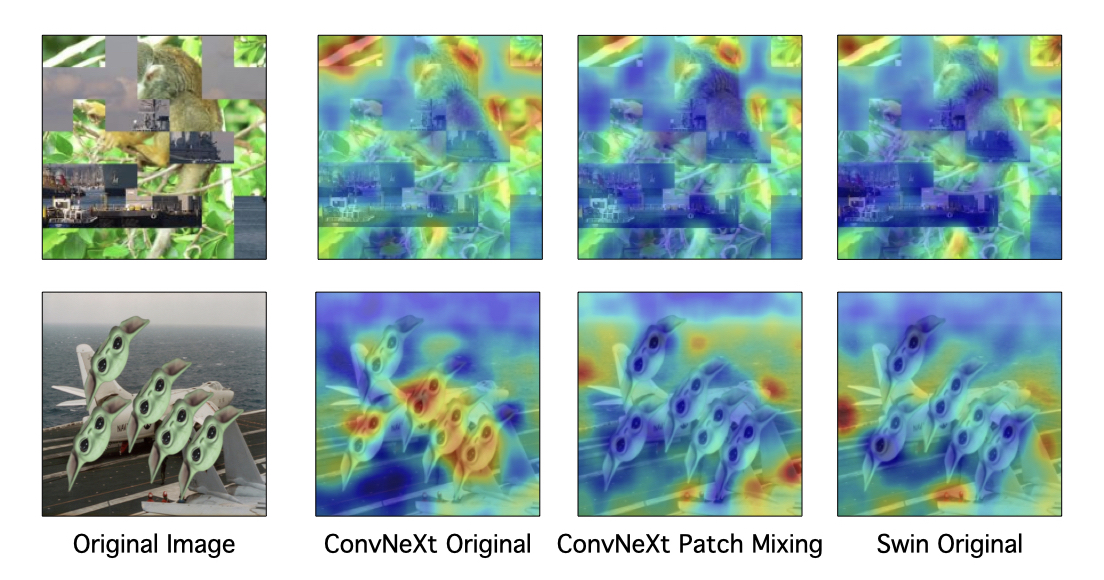}
    \caption{\centering Saliency maps of spider monkey (top) and airplane carrier (bottom). \textit{ConvNeXt w/ Patch Mixing shows a strongly improved ability to ignore out-of-context patches.}}
    \label{fig:saliency_maps}
\end{figure}

\paragraph{CNNs trained with Patch Mixing exhibit increased patch selectivity, rivaling that of ViTs} We show the quantitative results of inverse patch selectivity in Table \ref{c-rise table} for Tiny networks using grid sizes of (7, 7) and (14, 14). We also illustrate the differences between the models' heatmap appearances in Figure~\ref{fig:saliency_maps}. Specifically, we can see how ConvNeXt Original's importance map \textit{spills} from in-context to out-of-context patches due to the convolutional architecture, a phenomenon that is addressed in ConvNeXt w/ Patch Mixing. ConvNeXt Patch Mixing and Swin Original both correctly classify the airplane carrier in Figure~\ref{fig:saliency_maps}, but ConvNeXt original incorrectly classifies the image as carousel. This shows that ConvNeXt Patch Mixing more effectively ignores occluders that are out-of-context in general, with importance maps that mirror those of Swin.

\begin{table}
\caption{Inverse patch selectivity (\textbf{lower} is better) using c-RISE and patch attack grid sizes of (7, 7) and (14, 14). We evaluate 5 images per class for 100 classes using Softmax normalized saliency maps.} 
\label{c-rise table}
\centering
    \begin{tabular}{lcc}
    \toprule
    \multicolumn{1}{c}{Model} & \multicolumn{2}{c}{Inverse Patch Selectivity} \\ 
    \cmidrule(lr){2-3}
    & (7, 7) & (14, 14) \\
    \toprule
    ConvNeXt-T Original & 0.0201 & 0.0198\\ 
    ConvNeXt-T Patch Mixing & \textbf{0.0194} & \textbf{0.0196}\\ 
    \midrule
    Swin-T Original & 0.0196 & 0.0197 \\ 
    Swin-T Patch Mixing & 0.0197 & 0.0198\\
    \bottomrule
    \end{tabular}
\end{table}

\section{Related Work}

\paragraph{Data Augmentation}
There are many data augmentations that attempt to address the issue of occlusion, from stochastic elimination of regions within training images to regional dropout~\cite{zhong_re, cutout2017, Xiao_TDAPNet}. To effectively address the limitations of traditional empirical risk minimization approaches in training deep neural networks, Zhang et al.~\cite{zhang2017mixup} introduced Mixup. A now widely utilized data augmentation technique, Mixup synthesizes new training instances by linearly interpolating between random image pairs and their respective labels. This approach encourages the model to produce smoother decision boundaries, leading to better generalization. As noted by Yun et al.~\cite{yun2019cutmix}, Mixup samples are locally ambiguous and unnatural, often confusing the model. To address this, Yun et al. presented CutMix, a regularization strategy for training robust classifiers with localizable features. CutMix combines the benefits of previous data augmentation techniques, such as Mixup and Cutout ~\cite{cutout2017}, by overlaying patches of one image onto another and adjusting the corresponding labels proportionally. 

\paragraph{Occlusion}
Current related works on occlusion in object detection and image classification indicate that while systems have evolved to be more robust, they still fail to accurately classify and detect objects under severe occlusion. Existing approaches like Region Proposal Networks~\cite{Girshick_region}, which are applied for learning fast detection approaches~\cite{Girshick_fast}, perform well for object detection tasks but fail when the bounding box of an object is occluded. Recent works have shown that traditional approaches like Deep Convolutional Neural Networks (DCNNs) such as ResNet~\cite{ResNet} or VGG~\cite{VGG} display little robustness to occlusion~\cite{zhu_occ, Liu_occ}. Addressing this issue with data augmentations simulating partial occlusion has had limited success~\cite{cutout2017}. Conversely, generative compositional models have been shown to be robust to partial object occlusion with the ability to still detect object features~\cite{ya_context, Fidler, dai_hong, wang_occ}. Recently, CompositionalNets, which incorporate DCNN architecture, have been proven to be far more robust to occlusion than their traditional counterparts ~\cite{Kortylewski_2020_CVPR, Kortylewski_greedy}. Building off this work, context-aware CompositionalNets were introduced to control the influence of the object's context on the classification result, increasing accuracy when confronted with largely occluded objects~\cite{wang_compnets}. Other deep learning approaches require detailed part-level annotations to reconstruct occluded objects, which is costly~\cite{zhang_occ_aware, onet_cvpr19}.

\section{Conclusion}

In this paper, we investigated the difference between CNNs and ViTs in terms of their ability to handle occlusion and ignore out-of-context information. In particular, we introduced the concept of \textit{patch selectivity} as a measure of this ability and showed that ViTs naturally possess higher patch selectivity than CNNs. We also proposed Patch Mixing, a data augmentation method that simulates patch selectivity in CNNs by inserting patches from other images onto training images. We demonstrated that Patch Mixing improves the performance of CNNs on various occlusion benchmarks, including two new datasets that we created: SMD and ROD. Furthermore, we developed c-RISE, a contrastive explainability method that allows us to visualize and quantify patch selectivity for both CNNs and ViTs. Our results suggest that patch selectivity is an important element for occlusion robustness and Patch Mixing is an effective method to amplify this characteristic within CNNs, bridging the gap with respect to ViTs that are naturally stronger in this area.

\bibliographystyle{abbrv}
\bibliography{neurips_2023}

\raggedbottom
\pagebreak

\section*{\LARGE Supplementary Material}

\section*{Superimposed Masked Dataset (SMD) Details}

\begin{figure}
    \centering
    \begin{subfigure}{.23\textwidth}
        \includegraphics[width=\linewidth, frame]{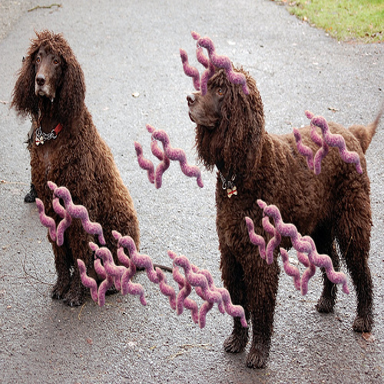}
    \end{subfigure}
    \begin{subfigure}{.23\textwidth}
        \includegraphics[width=\linewidth, frame]{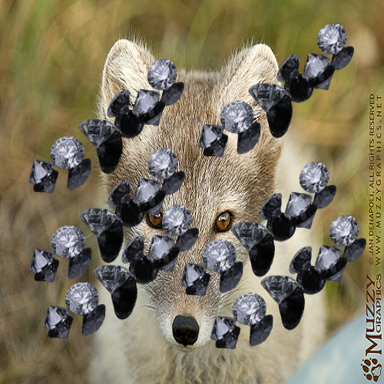}
    \end{subfigure}
    \begin{subfigure}{.23\textwidth}
        \includegraphics[width=\linewidth, frame]{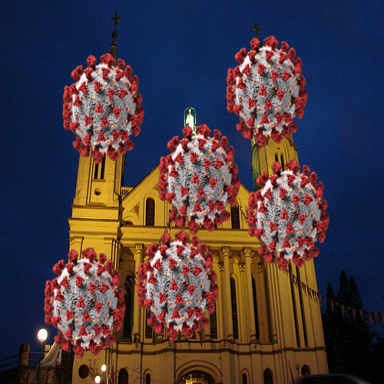}
    \end{subfigure}
    \begin{subfigure}{.23\textwidth}
        \includegraphics[width=\linewidth, frame]{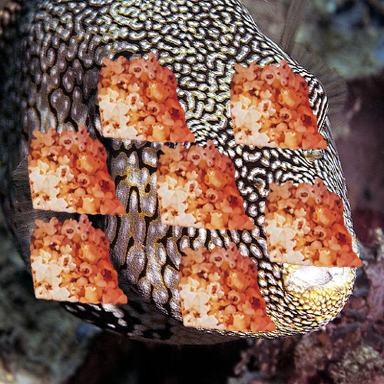}
    \end{subfigure}
    \begin{subfigure}{.23\textwidth}
        \includegraphics[width=\linewidth, frame]{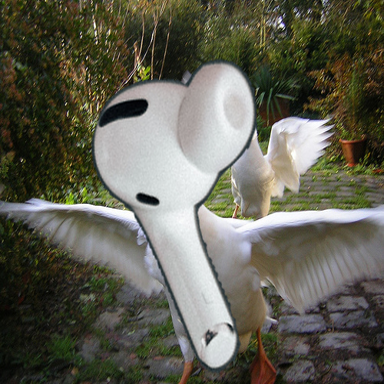}
    \end{subfigure}
    \begin{subfigure}{.23\textwidth}
        \includegraphics[width=\linewidth, frame]{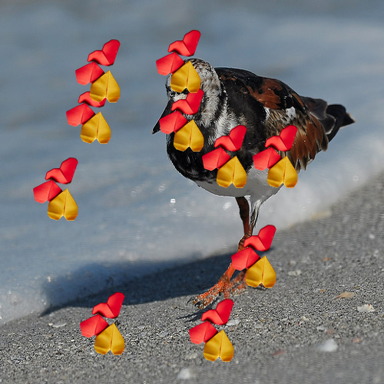}
    \end{subfigure}
    \begin{subfigure}{.23\textwidth}
        \includegraphics[width=\linewidth, frame]{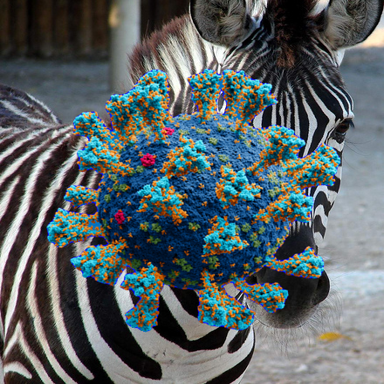}
    \end{subfigure}
    \begin{subfigure}{.23\textwidth}
        \includegraphics[width=\linewidth, frame]{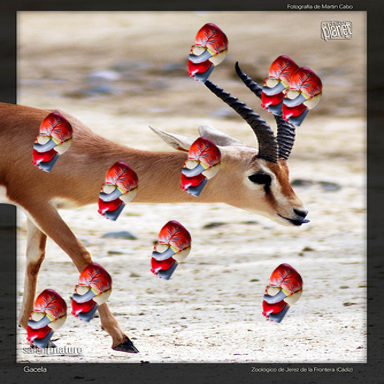}
    \end{subfigure}
    \begin{subfigure}{.23\textwidth}
        \includegraphics[width=\linewidth, frame]{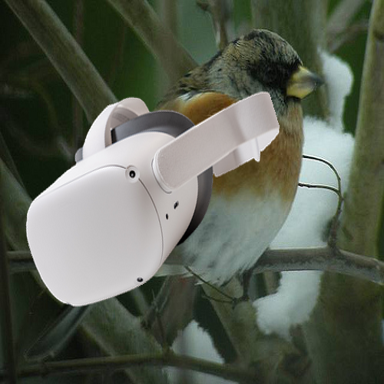}
    \end{subfigure}
    \begin{subfigure}{.23\textwidth}
        \includegraphics[width=\linewidth, frame]{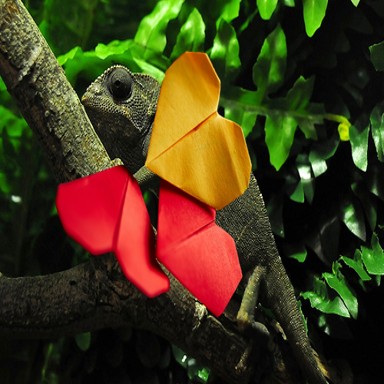}
    \end{subfigure}
    \begin{subfigure}{.23\textwidth}
        \includegraphics[width=\linewidth, frame]{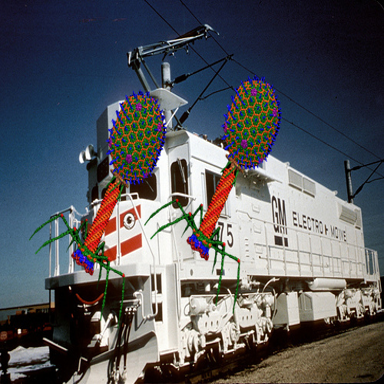}
    \end{subfigure}
    \begin{subfigure}{.23\textwidth}
        \includegraphics[width=\linewidth, frame]{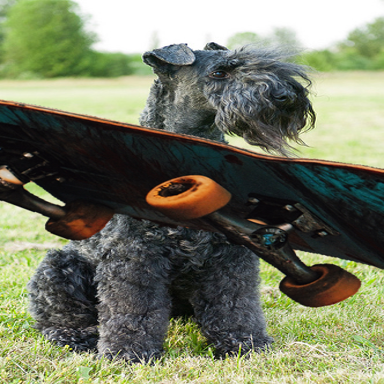}
    \end{subfigure}
    \begin{subfigure}{.23\textwidth}
        \includegraphics[width=\linewidth, frame]{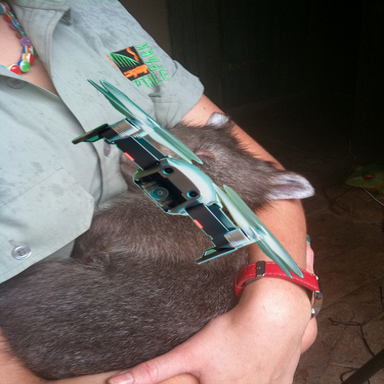}
    \end{subfigure}
    \begin{subfigure}{.23\textwidth}
        \includegraphics[width=\linewidth, frame]{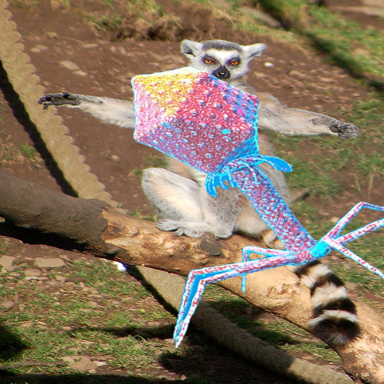}
    \end{subfigure}
    \begin{subfigure}{.23\textwidth}
        \includegraphics[width=\linewidth, frame]{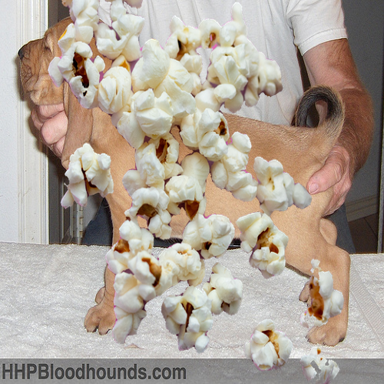}
    \end{subfigure}
    \begin{subfigure}{.23\textwidth}
        \includegraphics[width=\linewidth, frame]{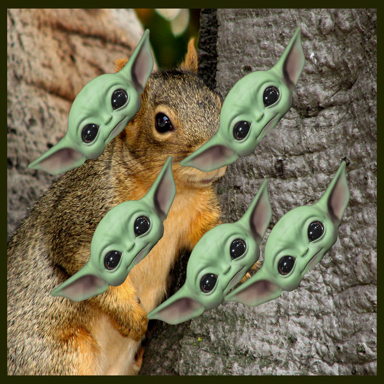}
    \end{subfigure}
    \caption{Examples from SMD with occlusion levels between 10-30\%.}
    \label{fig:SMD}
\end{figure}

\begin{figure}[H]
    \centering
    \fbox{\begin{subfigure}{.12\textwidth}
        \includegraphics[width=\linewidth]{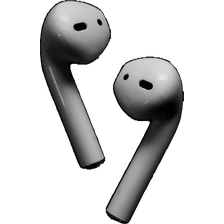}
    \end{subfigure}}
    \fbox{\begin{subfigure}{.12\textwidth}
        \includegraphics[width=\linewidth]{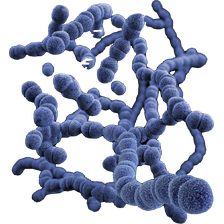}
    \end{subfigure}}
    \fbox{\begin{subfigure}{.12\textwidth}
        \includegraphics[width=\linewidth]{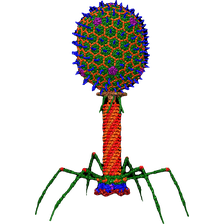}
    \end{subfigure}}
    \fbox{\begin{subfigure}{.12\textwidth}
        \includegraphics[width=\linewidth]{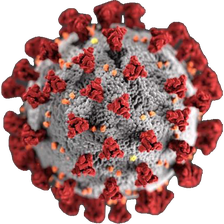}
    \end{subfigure}}
    \fbox{\begin{subfigure}{.12\textwidth}
        \includegraphics[width=\linewidth]{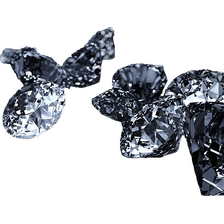}
    \end{subfigure}}
    \fbox{\begin{subfigure}{.12\textwidth}
        \includegraphics[width=\linewidth]{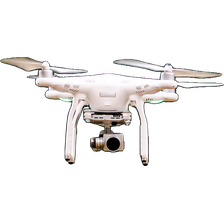}
    \end{subfigure}}
    \fbox{\begin{subfigure}{.12\textwidth}
        \includegraphics[width=\linewidth]{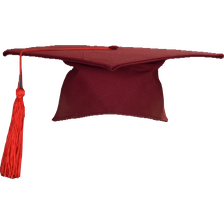}
    \end{subfigure}}\\
    \fbox{\begin{subfigure}{.12\textwidth}
        \includegraphics[width=\linewidth]{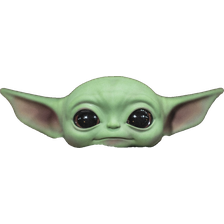}
    \end{subfigure}}
    \fbox{\begin{subfigure}{.12\textwidth}
        \includegraphics[width=\linewidth]{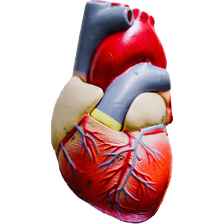}
    \end{subfigure}}
    \fbox{\begin{subfigure}{.12\textwidth}
        \includegraphics[width=\linewidth]{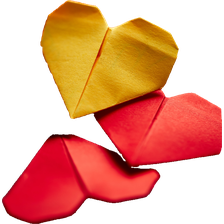}
    \end{subfigure}}
    \fbox{\begin{subfigure}{.12\textwidth}
        \includegraphics[width=\linewidth]{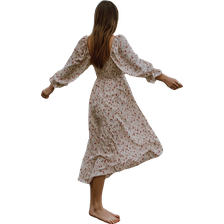}
    \end{subfigure}}
    \fbox{\begin{subfigure}{.12\textwidth}
        \includegraphics[width=\linewidth]{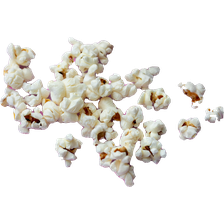}
    \end{subfigure}}
    \fbox{\begin{subfigure}{.12\textwidth}
        \includegraphics[width=\linewidth]{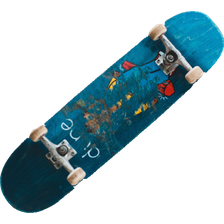}
    \end{subfigure}}
    \fbox{\begin{subfigure}{.12\textwidth}
        \includegraphics[width=\linewidth]{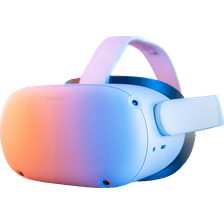}
    \end{subfigure}}
    \caption{One example for each class of occluder object in SMD. From left to right, the occluders in the first row are: \textit{airpods, bacteria, bacteriophage, coronavirus, diamonds, drone, and graduation cap}. Similarly for the second row: \textit{Grogu (baby yoda), anatomical heart, origami heart, person, popcorn, skateboard, and virtual reality headset.}}
    \label{fig:occluders}
\end{figure}

Here we present additional details and experimental results regarding SMD, which is introduced in Section 3 of the main paper. Figure~\ref{fig:SMD} provides additional images from SMD, and Figure~\ref{fig:occluders} shows one example of each occluder for each class.

Occluder objects are randomly selected and rotated prior to being applied to the validation images. So as not to completely occlude important image features, we place multiple instances of the same occluder object on each image. Pertinent information, including occluder masks, classes, and percentage of occlusion, is saved for future use. For lower levels of occlusion, the occluders do not overlap. For images with higher levels of occlusion, overlapping occluders are taken into account when calculating the final percentage of occlusion. Occluders are added to the image until the desired level of occlusion is reached. Table \ref{summary_occlusion} provides a comparison of the performance of Tiny and Small networks' Top-1 accuracy on three different validation sets with occlusion levels of approximately 10\%, 20\%, and 30\%. For both Tiny and Small models, ConvNet Patch Mixing provides higher accuracy than the original model across the board. However, the Swin models are always superior to the ConvNeXt models, with Swin Patch Mixing outperforming or matching Swin Original everywhere except the 30\% and average SMD set using the Small networks.

\begin{table}[H]
\caption{Top-1 Accuracy on SMD. Three different validation sets of SMD are generated with occlusion levels of approximately 10\%, 20\%, and 30\%. The average of all datasets, which is reported in the main paper, is also included. ConvNeXt trained with Patch Mixing outperforms all original ConvNeXt networks.}
\label{summary_occlusion}
\centering
    \begin{tabular}{lcccccccc}
    \toprule
     \multicolumn{1}{c}{Architecture} & \multicolumn{4}{c}{Tiny} & \multicolumn{4}{c}{Small} \\ 
     \cmidrule(lr){2-5} \cmidrule(lr){6-9}
     & 10\% & 20\% & 30\% & Avg & 10\% & 20\% & 30\% & Avg.\\
     \toprule
     ConvNeXt Original & 63.2 & 41.1 & 38.6 & 47.6 & 65.1 & 42.8 & 40.4 & 49.4\\ 
     ConvNeXt Patch Mixing & \textbf{64.2} & \textbf{42.5} & \textbf{40.1} & \textbf{48.9} & \textbf{65.4} & \textbf{43.2} & \textbf{41.6} & \textbf{50.1}\\ 
     \midrule
     Swin Original & 68.0 & 51.5 & 49.9 & 56.5 & 71.0 & 55.5 & \textbf{54.8} &\textbf{60.4}\\ 
     Swin Patch Mixing & \textbf{68.5} & \textbf{52.9} & \textbf{50.1} & \textbf{57.2} & 71.0 & \textbf{55.8}& 53.8 & 60.2\\ 
     \bottomrule
     \end{tabular}
\end{table}

\section*{Realistic Occlusion Dataset (ROD) Details}

\begin{figure}
    \centering
    \begin{subfigure}{.23\textwidth}
        \includegraphics[width=\linewidth, frame]{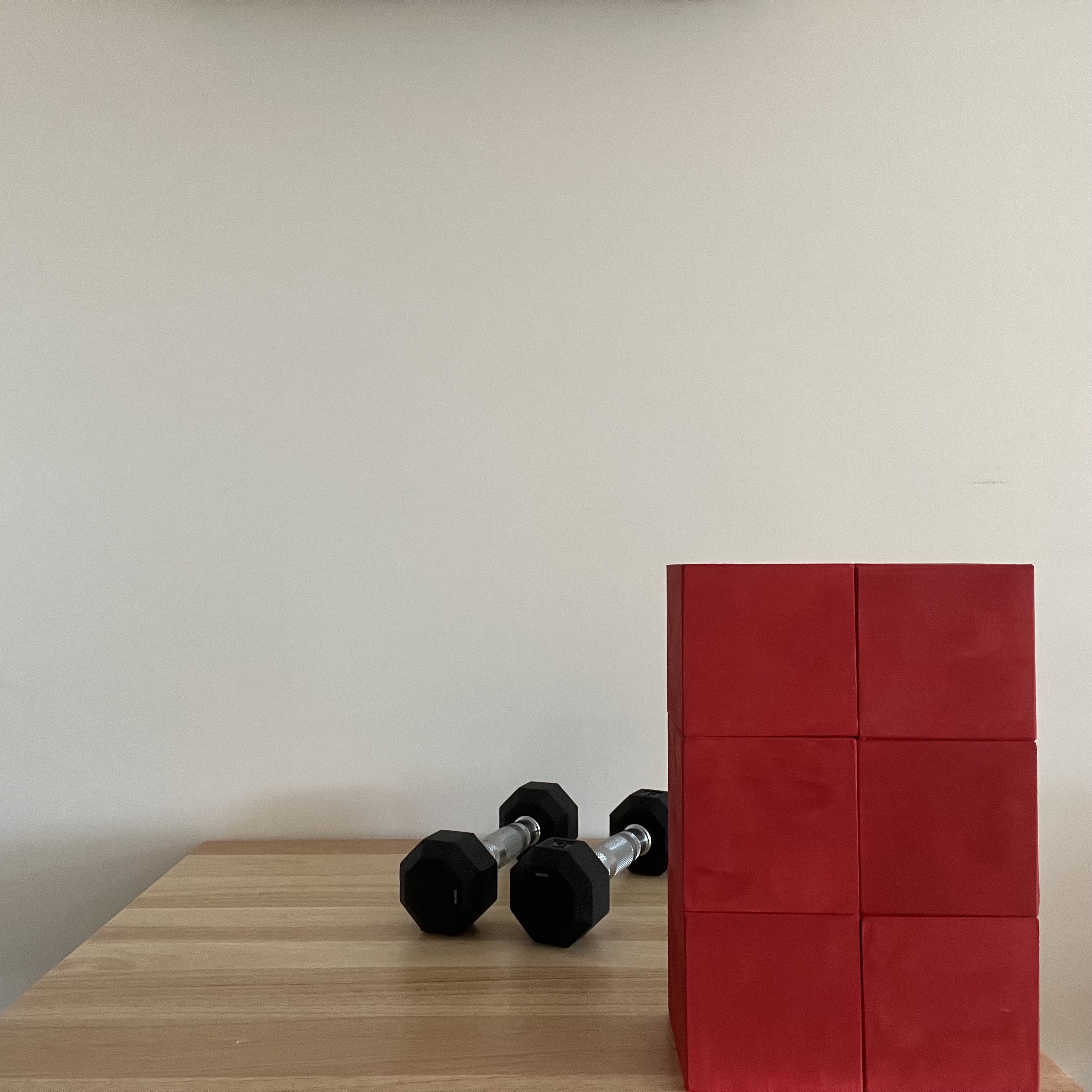}
    \end{subfigure}
    \begin{subfigure}{.23\textwidth}
        \includegraphics[width=\linewidth, frame]{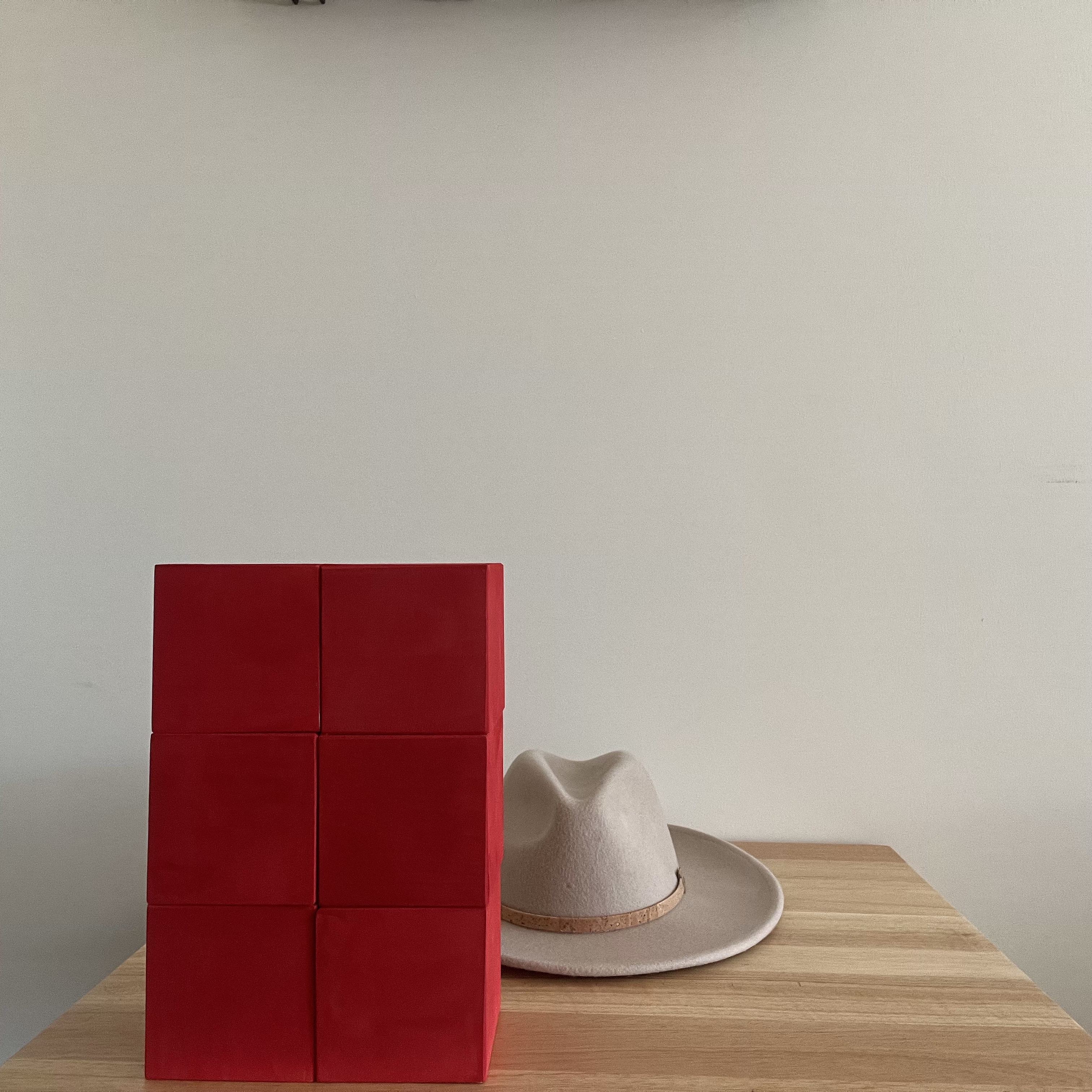}
    \end{subfigure}
    \begin{subfigure}{.23\textwidth}
        \includegraphics[width=\linewidth, frame]{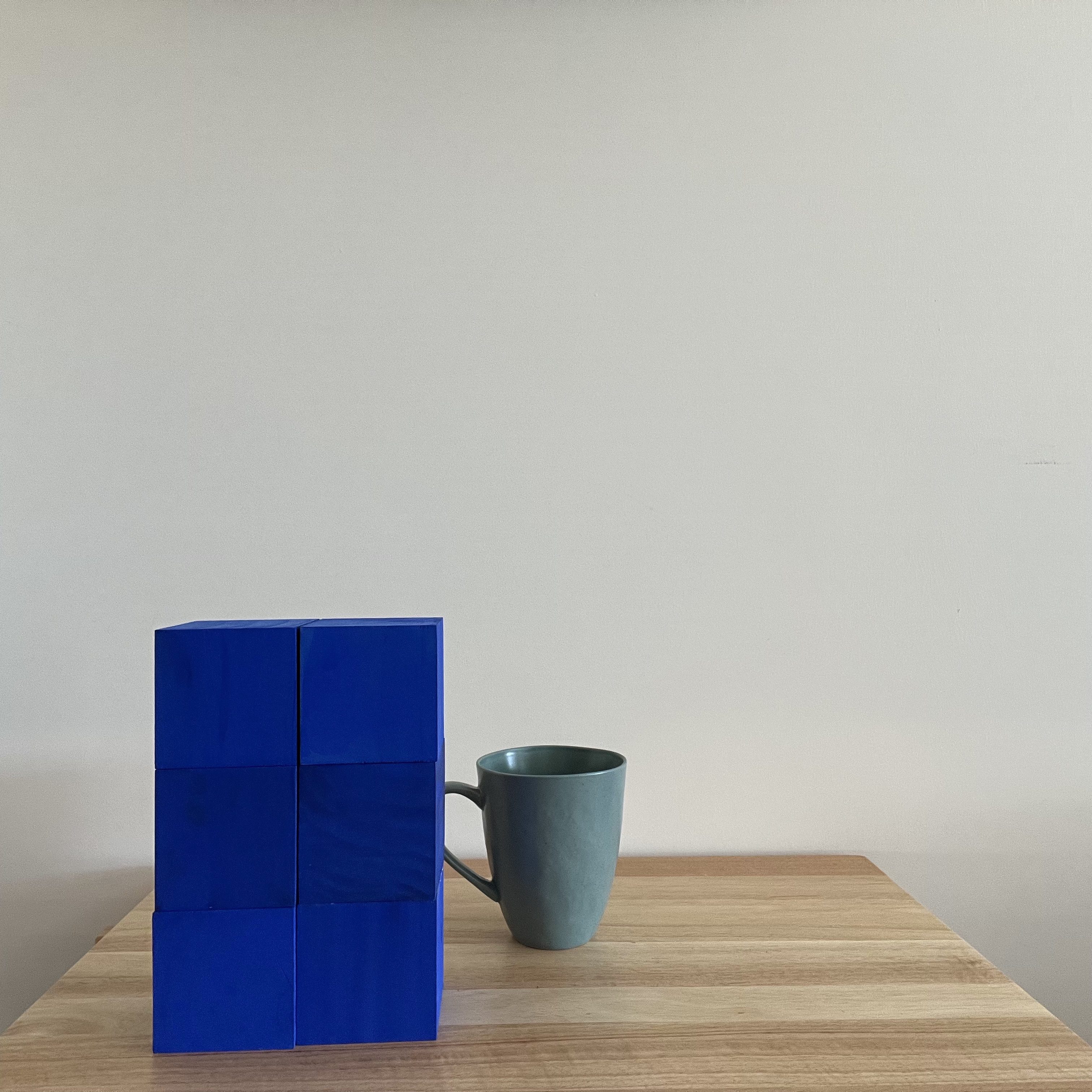}
    \end{subfigure}
    \begin{subfigure}{.23\textwidth}
        \includegraphics[width=\linewidth, frame]{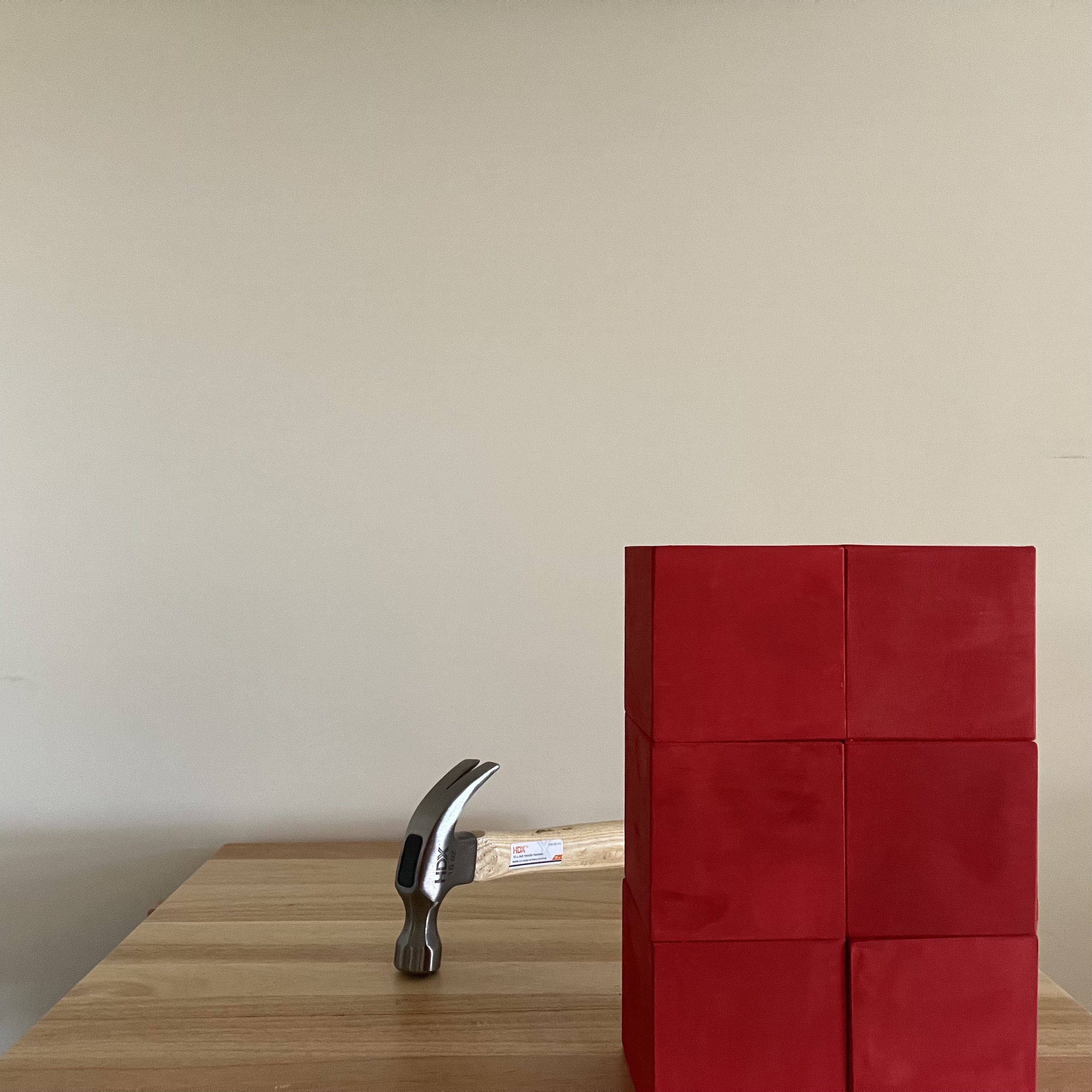}
    \end{subfigure}
    \begin{subfigure}{.23\textwidth}
        \includegraphics[width=\linewidth, frame]{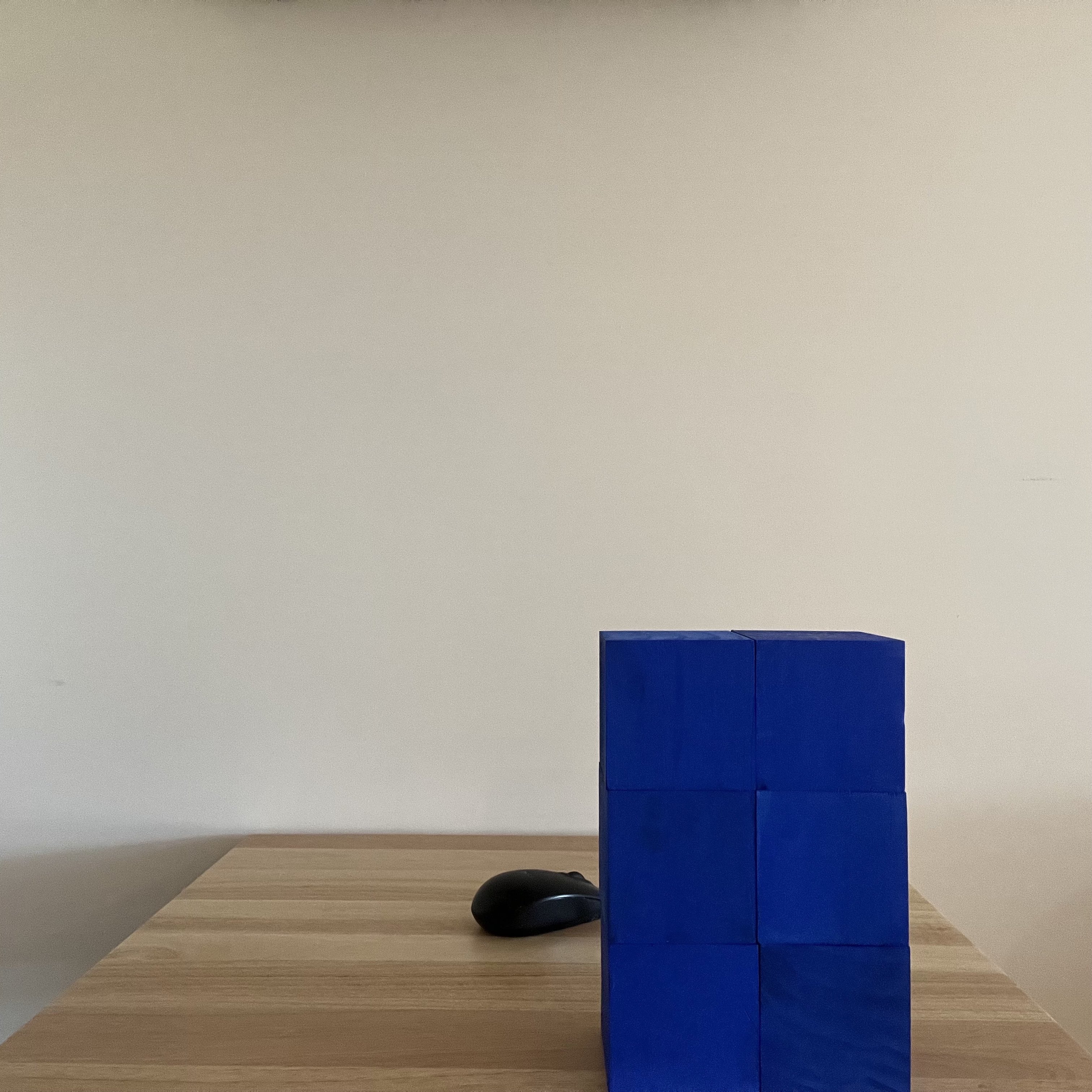}
    \end{subfigure}
    \begin{subfigure}{.23\textwidth}
        \includegraphics[width=\linewidth, frame]{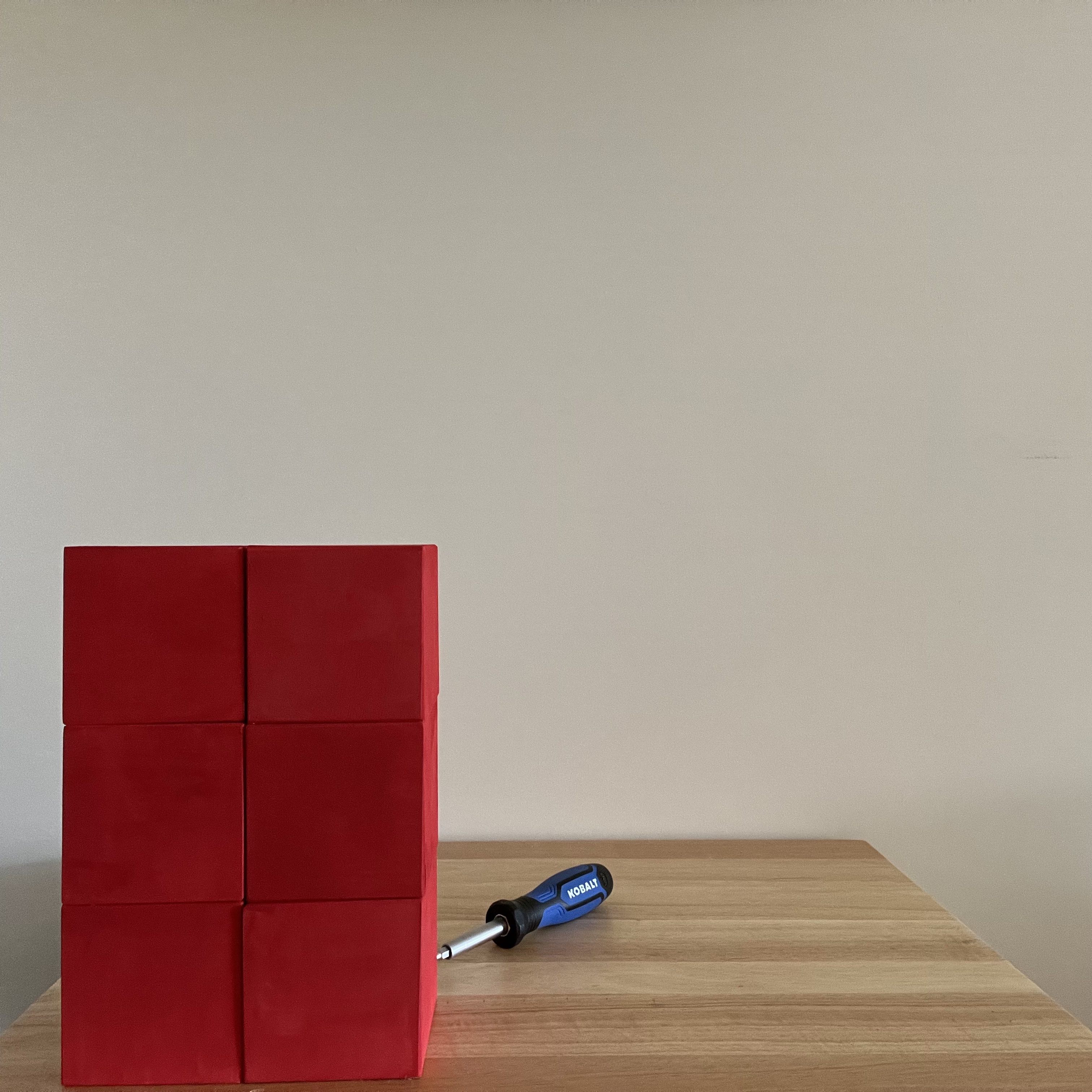}
    \end{subfigure}
    \begin{subfigure}{.23\textwidth}
        \includegraphics[width=\linewidth, frame]{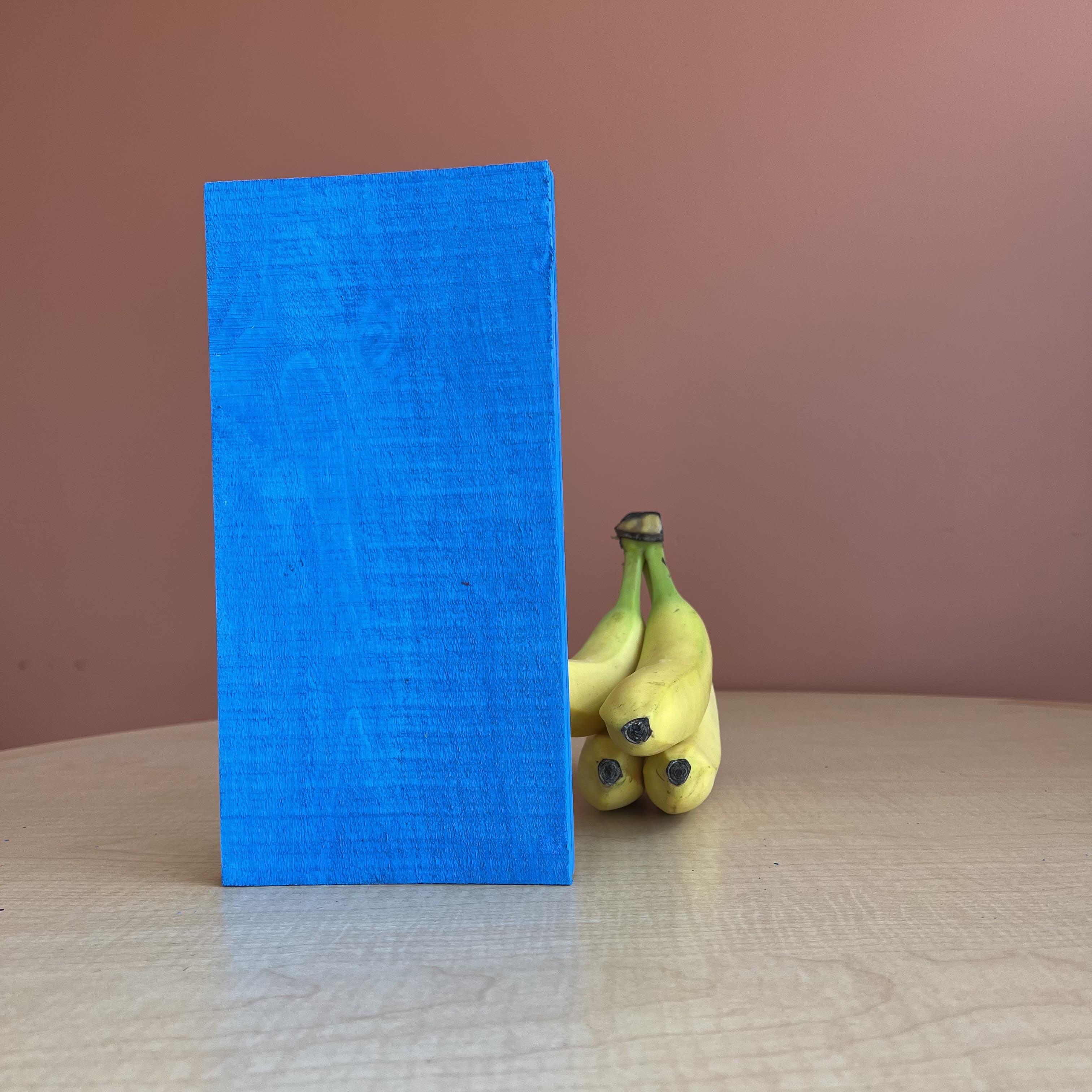}
    \end{subfigure}
    \begin{subfigure}{.23\textwidth}
        \includegraphics[width=\linewidth, frame]{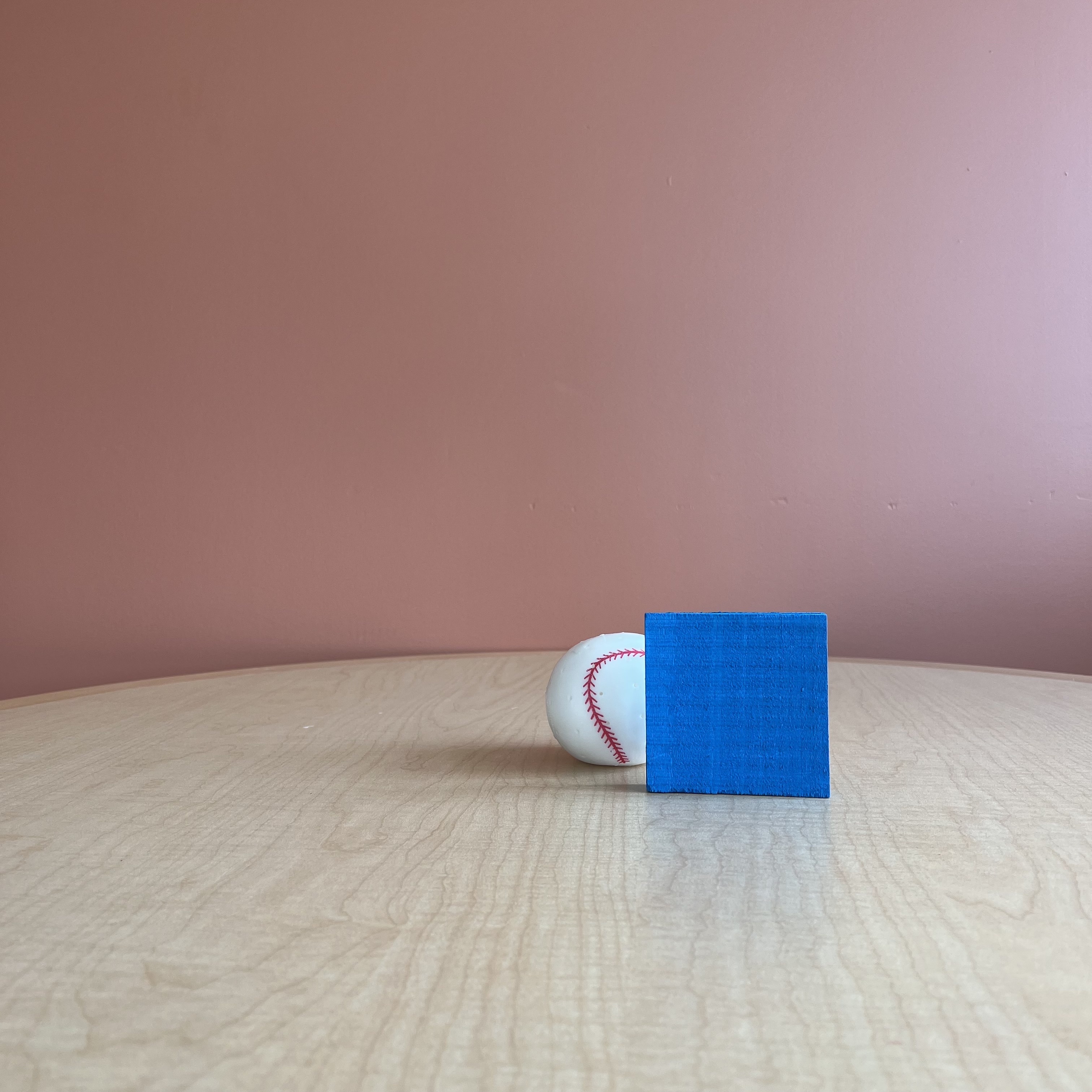}
    \end{subfigure}
    \begin{subfigure}{.23\textwidth}
        \includegraphics[width=\linewidth, frame]{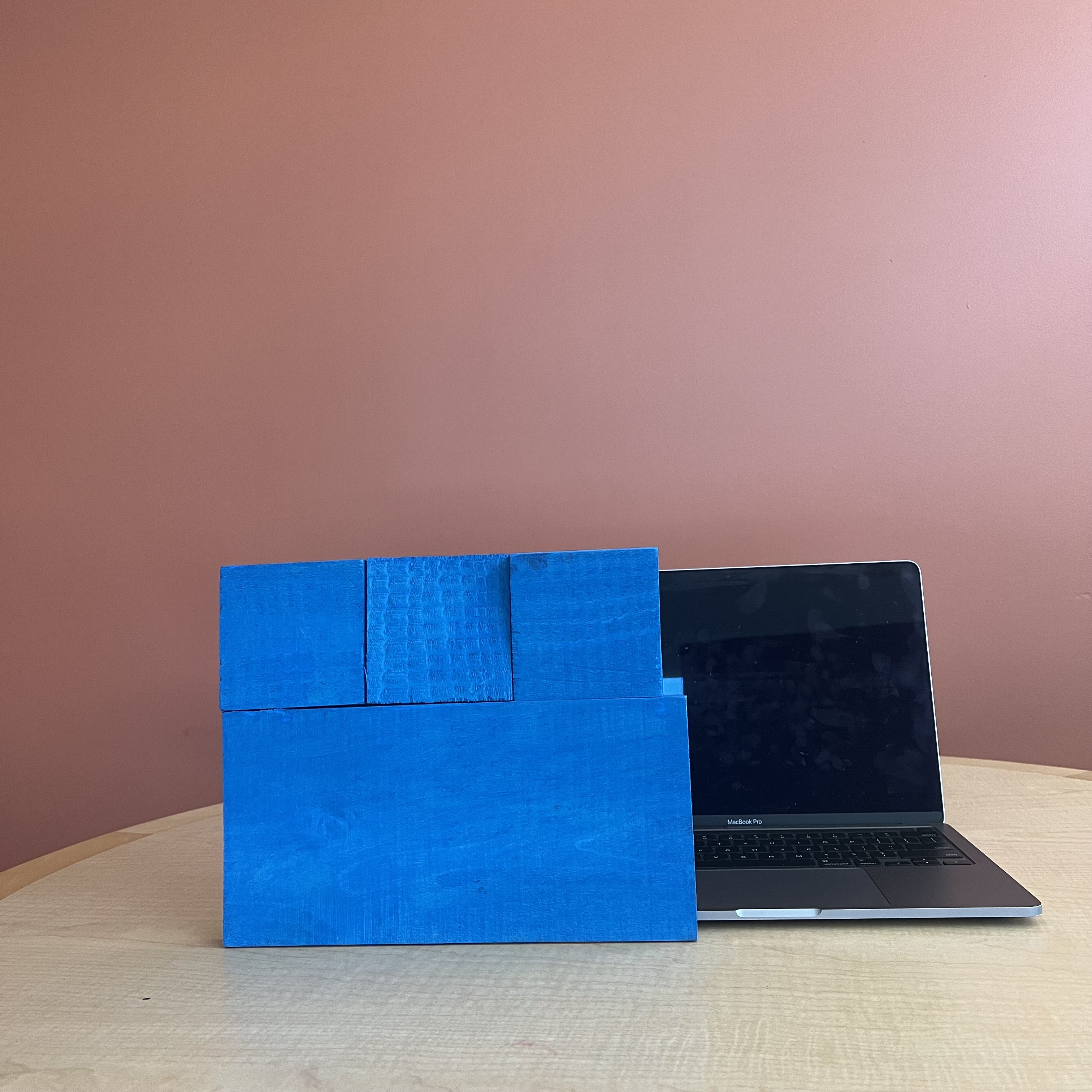}
    \end{subfigure}
    \begin{subfigure}{.23\textwidth}
        \includegraphics[width=\linewidth, frame]{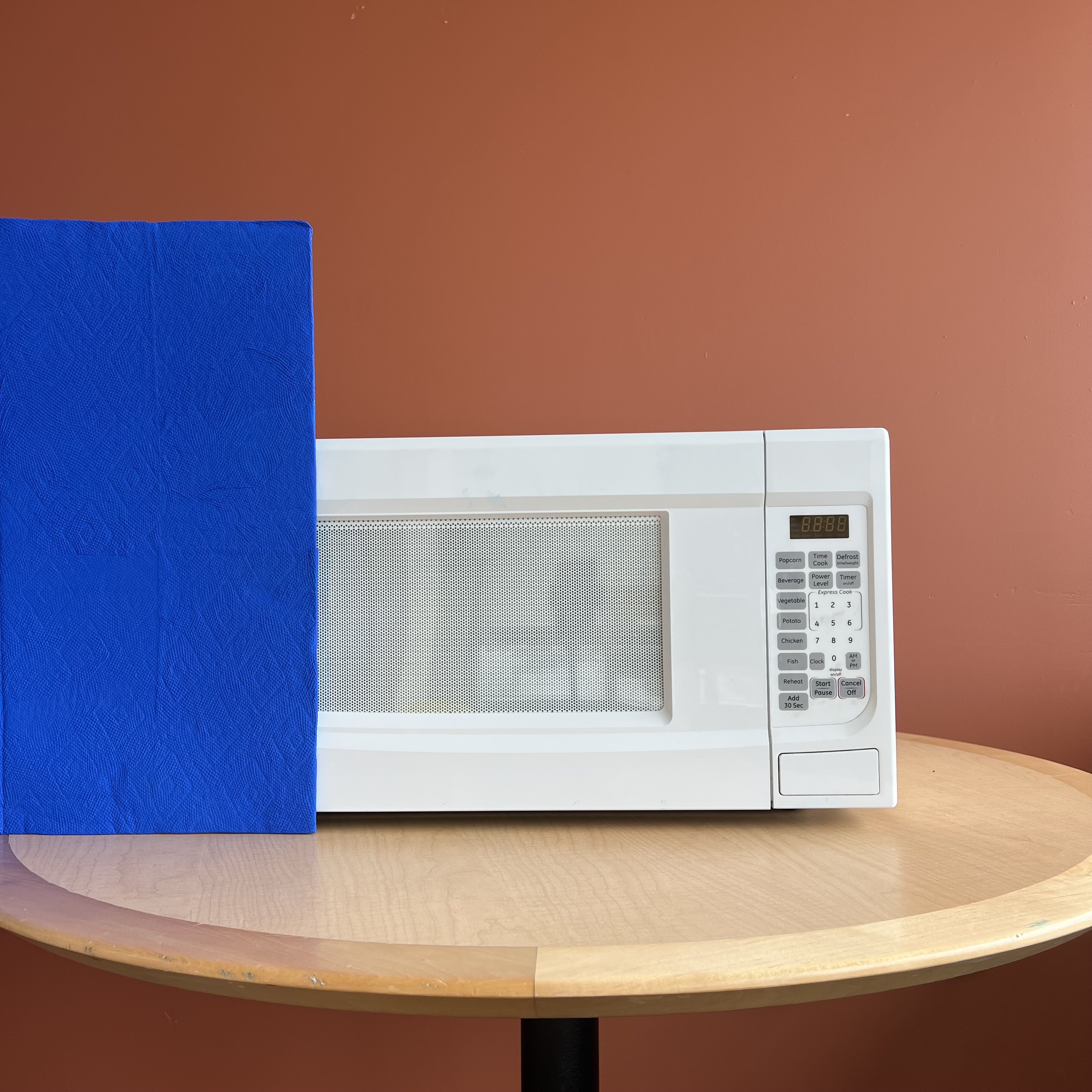}
    \end{subfigure}
    \begin{subfigure}{.23\textwidth}
        \includegraphics[width=\linewidth, frame]{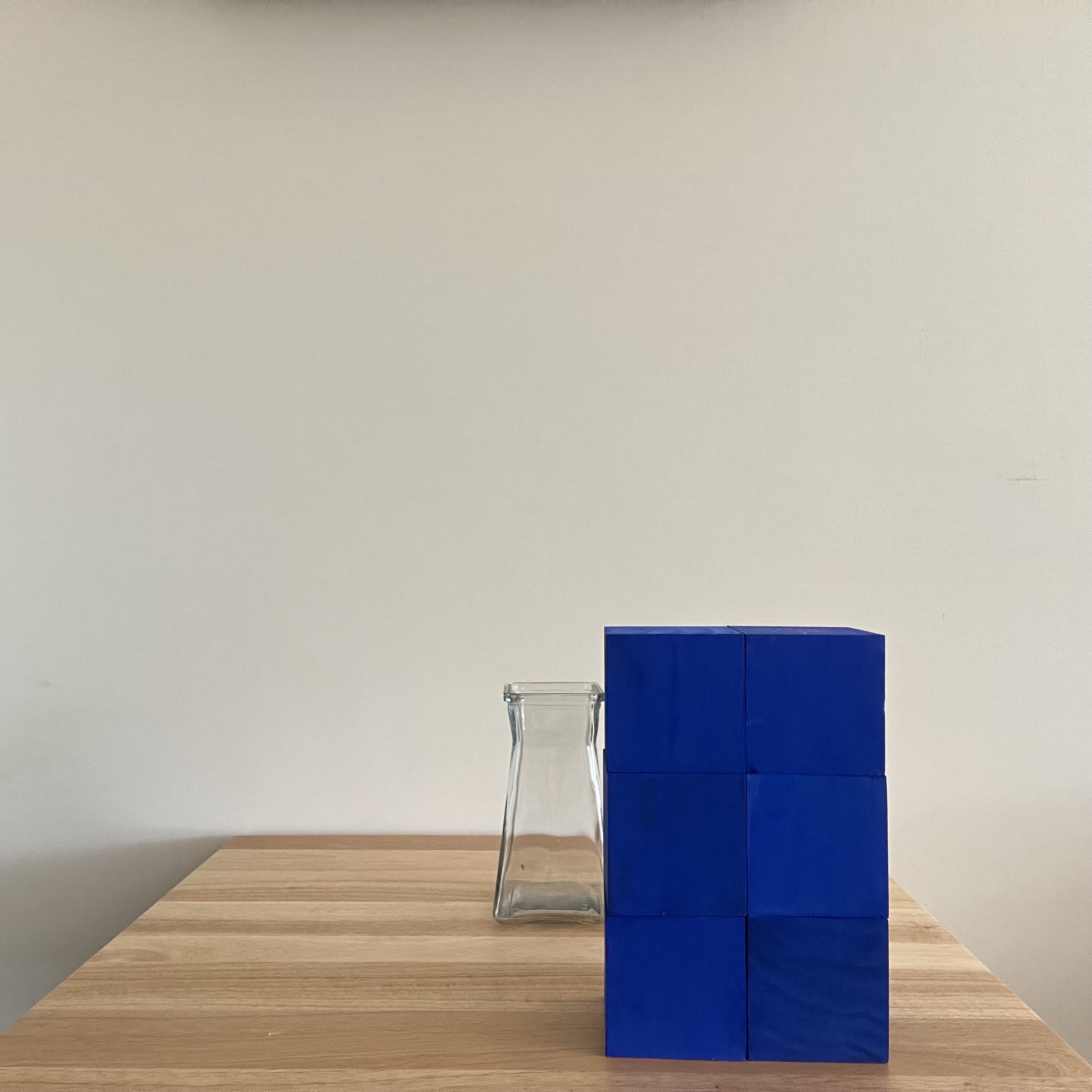}
    \end{subfigure}
    \begin{subfigure}{.23\textwidth}
        \includegraphics[width=\linewidth, frame]{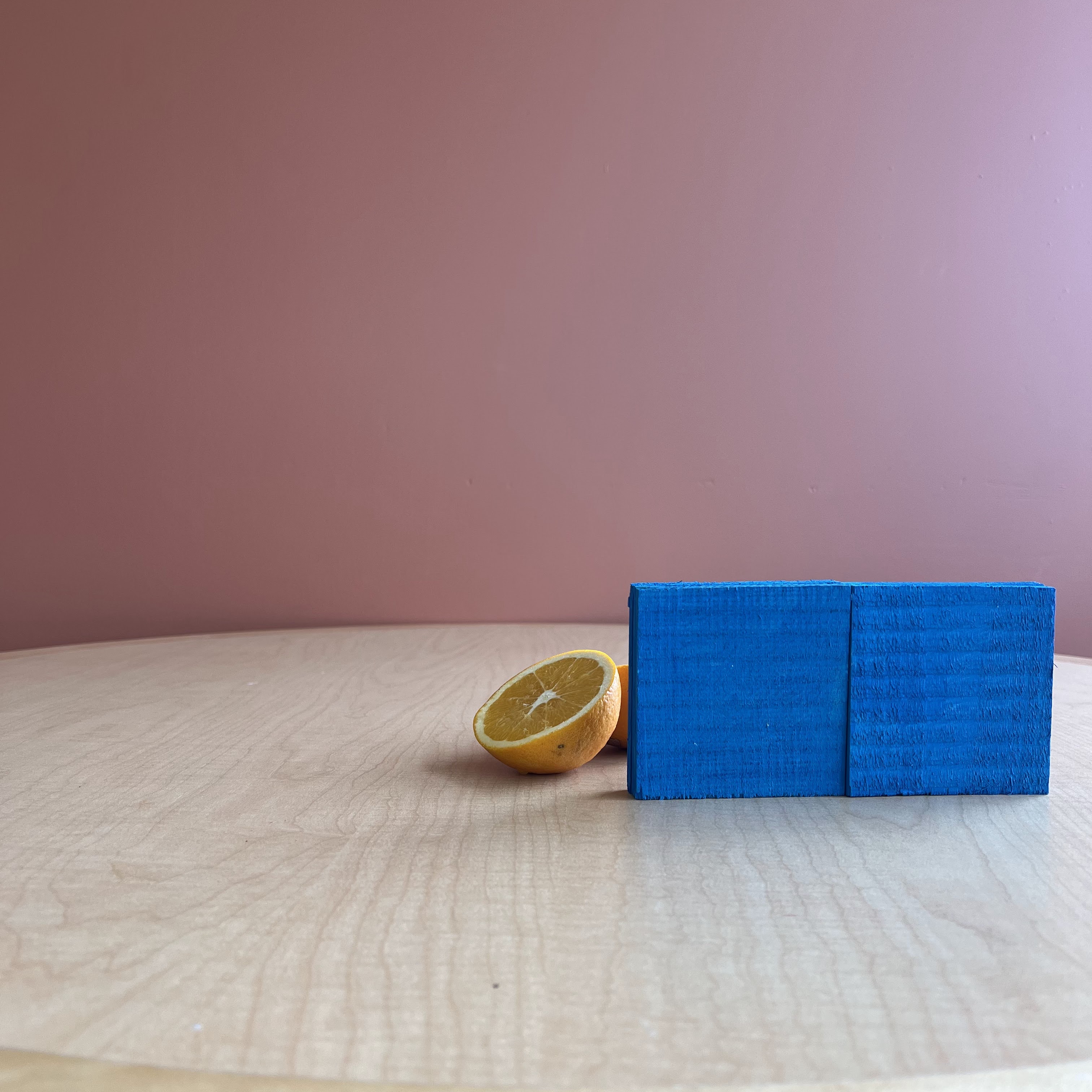}
    \end{subfigure}
    \begin{subfigure}{.23\textwidth}
        \includegraphics[width=\linewidth, frame]{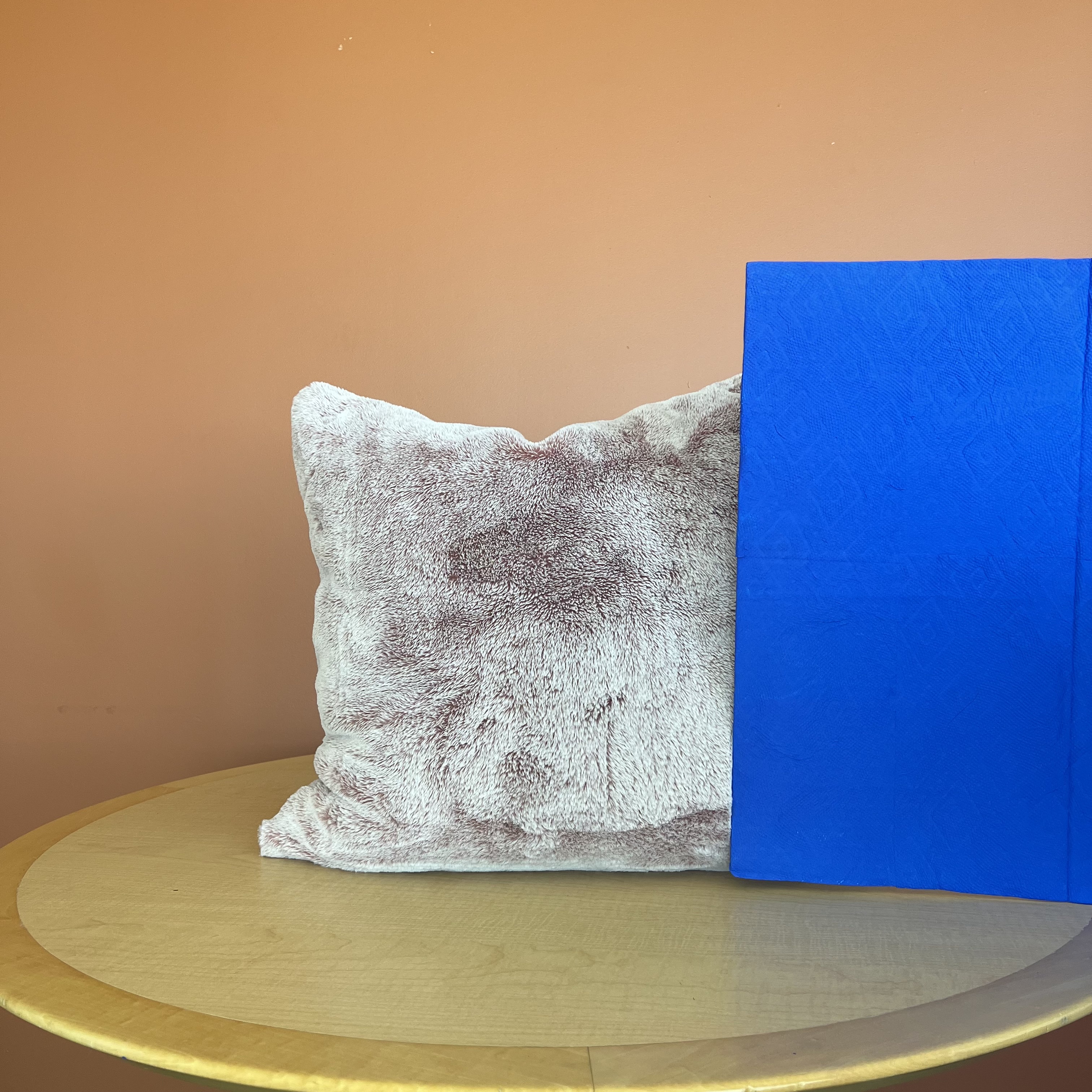}
    \end{subfigure}
    \begin{subfigure}{.23\textwidth}
        \includegraphics[width=\linewidth, frame]{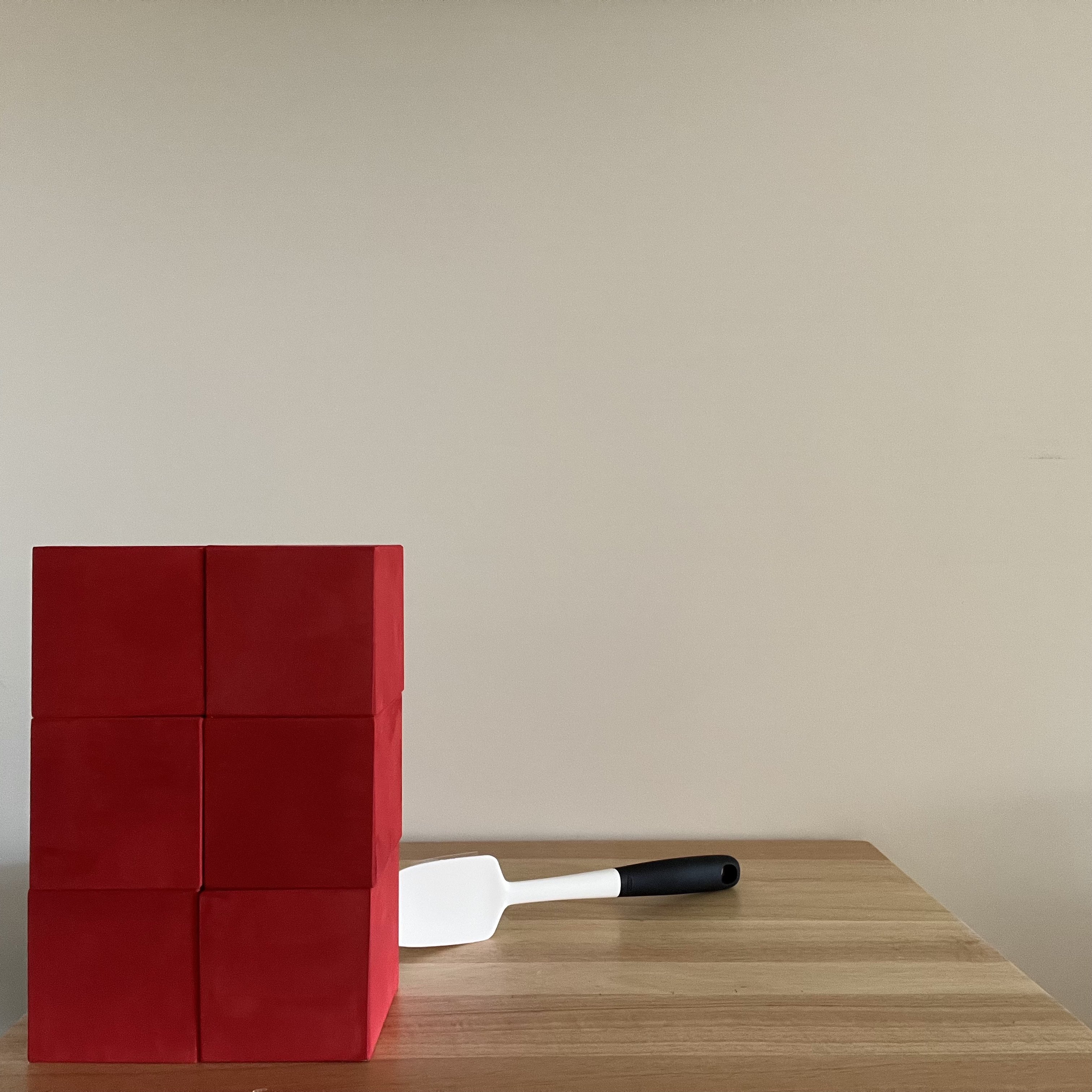}
    \end{subfigure}
    \begin{subfigure}{.23\textwidth}
        \includegraphics[width=\linewidth, frame]{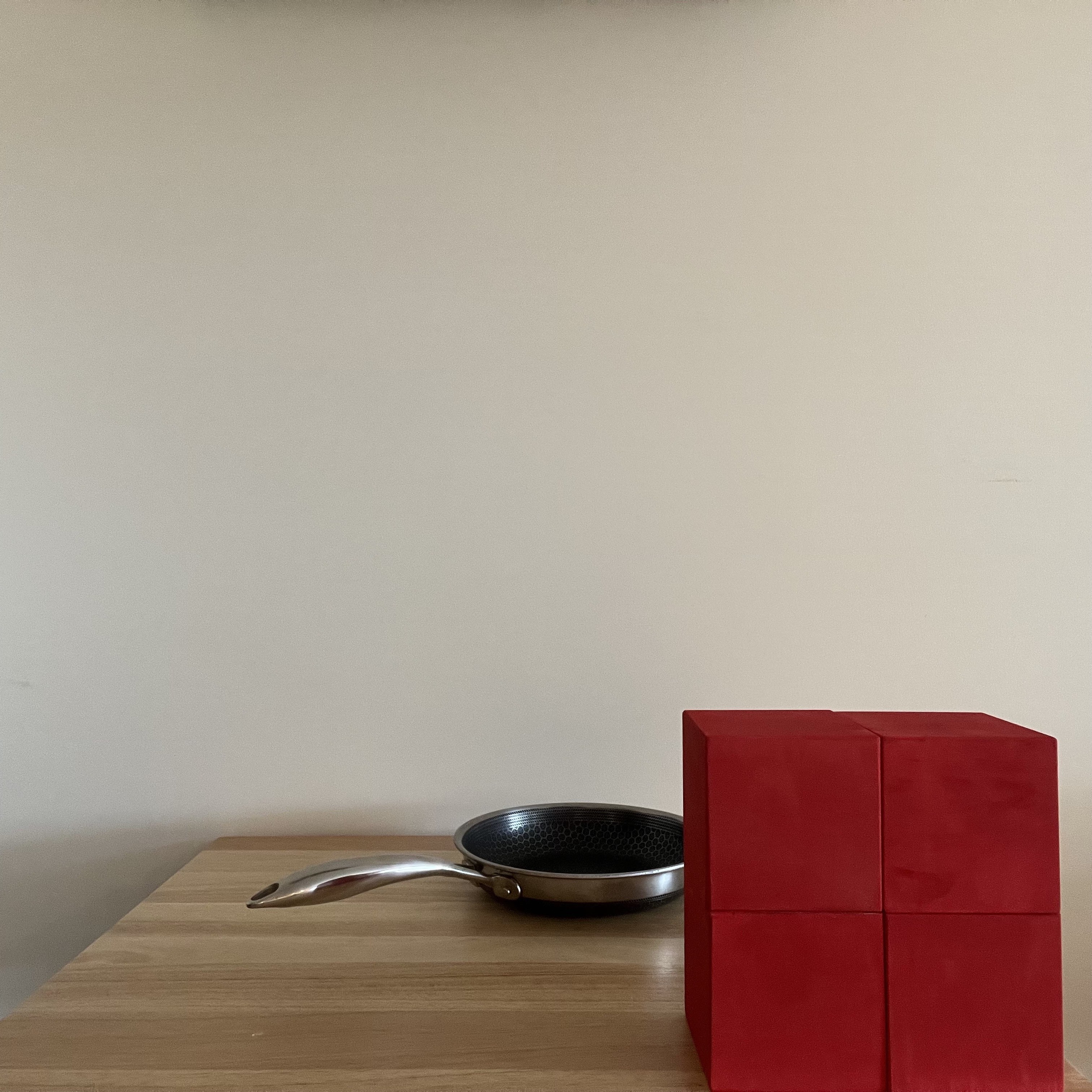}
    \end{subfigure}
    \begin{subfigure}{.23\textwidth}
        \includegraphics[width=\linewidth, frame]{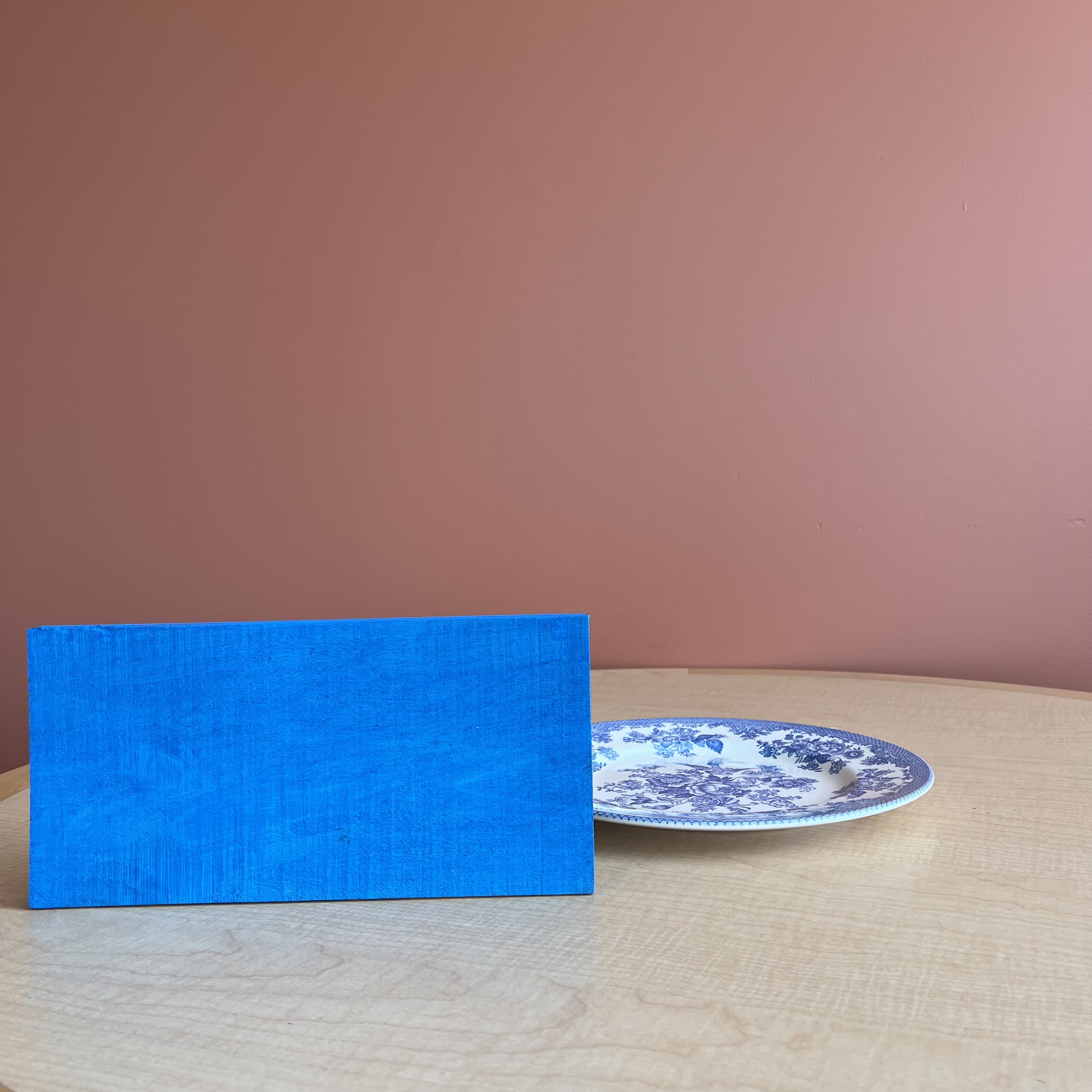}
    \end{subfigure}
    \caption{From left to right, Row 1: \textit{dumbbell, cowboy hat, cup, and hammer}. Row 2: \textit{mouse, screwdriver, banana, and baseball}. Row 3: \textit{laptop, microwave, vase, and orange}. Row 4: \textit{pillow, spatula, skillet, and plate.}}
    \label{fig:ROD}
\end{figure}

Figure~\ref{fig:ROD} contains representative samples of all 16 classes found in the ROD dataset, as elaborated in Section 3 of the main text. It's worth noting that this figure is not comprehensive, as ROD contains over 40 distinct objects. ConvNeXt-Tiny, when trained with Patch Mixing, outperforms the original model on ROD, while the performance of Small networks remains unaffected.

\section*{Naturalistic Variation Object Dataset (NVD)}

Counterfactual simulation testing of ConvNeXt and Swin networks was performed using the NVD occlusion subset~\cite{ruiz2022finding}. For this experiment, all occluded scenes were compared to an initial, non-occluded scene where the occluder was absent.

\begin{figure}
\centering
\includegraphics[width=0.24\linewidth]{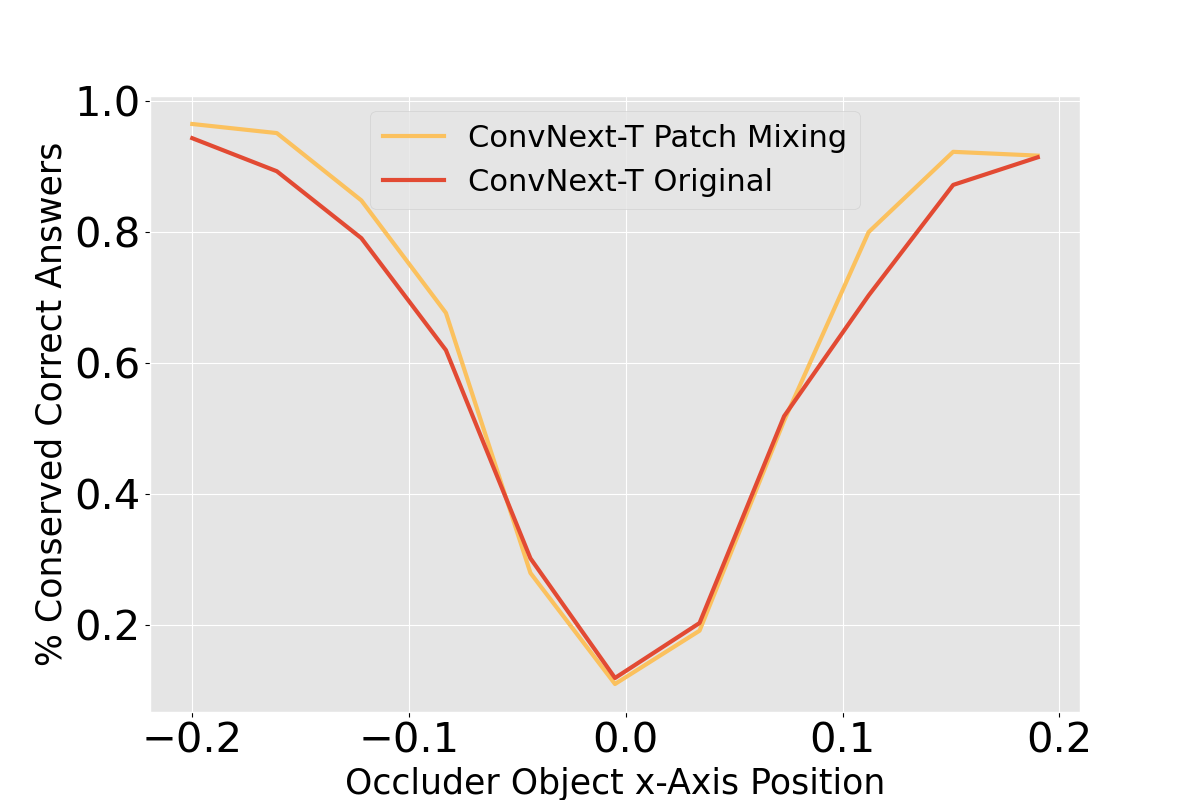}
\includegraphics[width=0.24\linewidth]{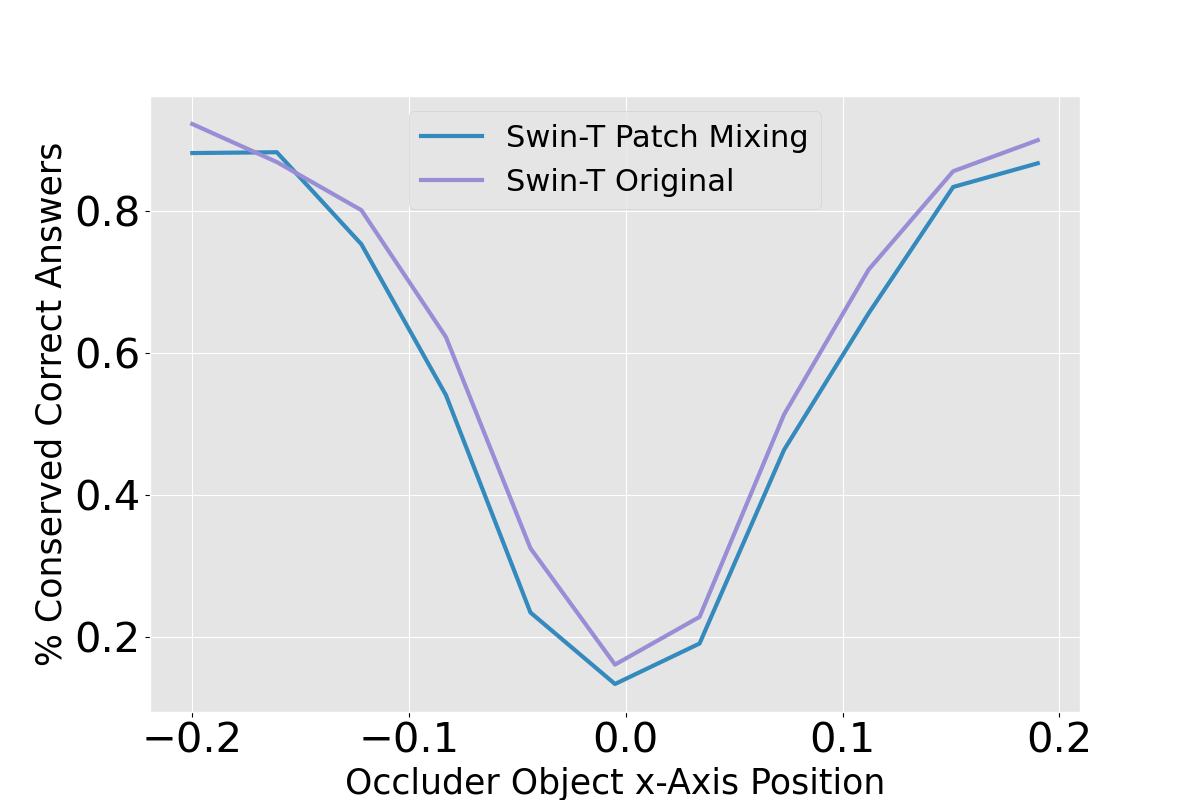}
\includegraphics[width=0.24\linewidth]{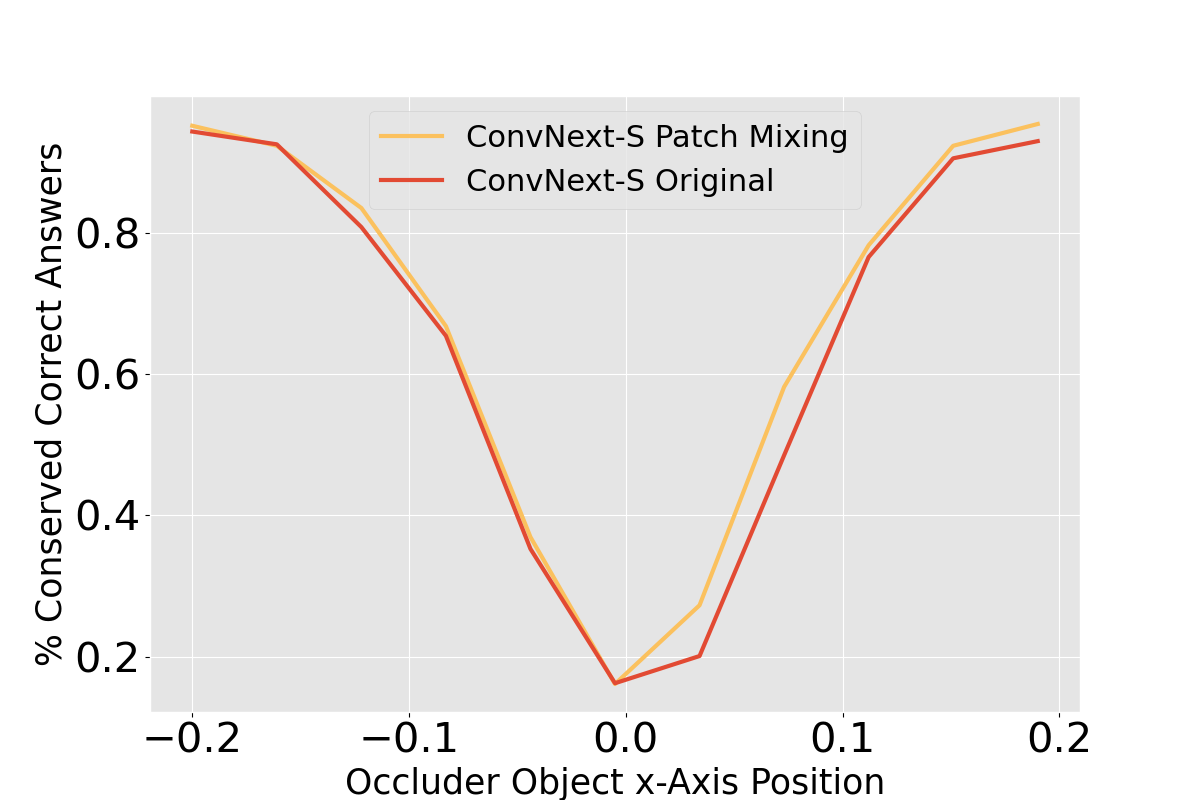}
\includegraphics[width=0.24\linewidth]{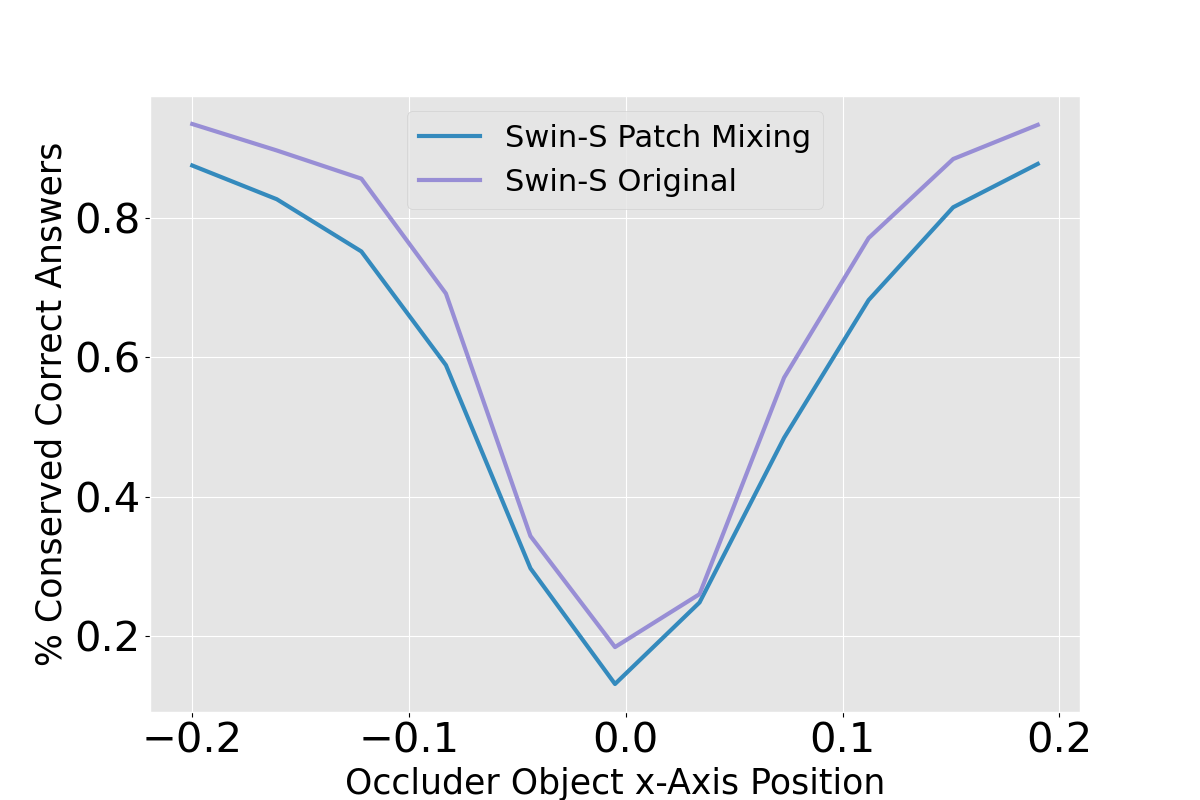}
\caption{ Occluder object x-axis position: Tiny and Small networks (NVD)}
\label{fig:axis}
\end{figure}

As seen in Figure~\ref{fig:axis}, both Swin and ConvNeXt performances for Tiny and Small networks significantly decline as the main object is increasingly occluded, reaching a minimum at zero on the x-axis. Notwithstanding, ConvNeXt networks trained with Patch Mixing demonstrate enhanced occlusion robustness compared to their original versions. For less substantial occlusions, the ConvNeXt Patch Mixing networks outperform the original models and even the Swin models. This trend is more significant in the Tiny networks, although unlike the small networks performance deteriorates and falls below Swin original as the occluder approaches the center of the main object. Interestingly, the results of Swin Patch Mixing are much worse than the originals for Tiny and Small networks regardless of the occluder's position.
\section*{Patch Mixing}
\begin{figure}
    \centering
    \begin{subfigure}{.15\textwidth}
        \centering
        \includegraphics[width=\linewidth, frame]{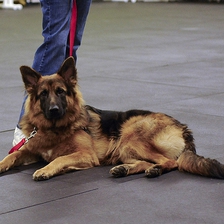}
    \end{subfigure}
    \hfill
    \begin{subfigure}{.15\textwidth}
        \centering
        \includegraphics[width=\linewidth, frame]{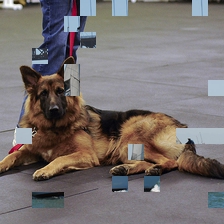}
    \end{subfigure}
    \begin{subfigure}{.15\textwidth}
        \centering
        \includegraphics[width=\linewidth, frame]{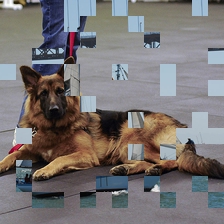}
    \end{subfigure}
    \begin{subfigure}{.15\textwidth}
        \centering
        \includegraphics[width=\linewidth, frame]{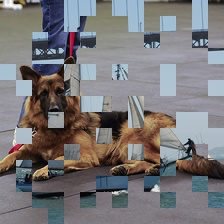}
    \end{subfigure}
    \begin{subfigure}{.15\textwidth}
        \centering
        \includegraphics[width=\linewidth, frame]{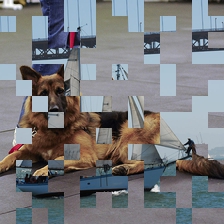}
    \end{subfigure}
    \begin{subfigure}{.15\textwidth}
        \centering
        \includegraphics[width=\linewidth, frame]{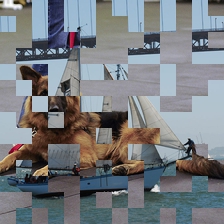}
    \end{subfigure} \\
    \begin{subfigure}{.15\textwidth}
        \centering
        \includegraphics[width=\linewidth, frame]{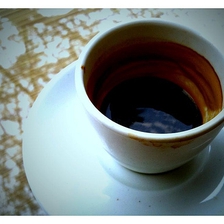}
    \end{subfigure}
    \hfill
    \begin{subfigure}{.15\textwidth}
        \centering
        \includegraphics[width=\linewidth, frame]{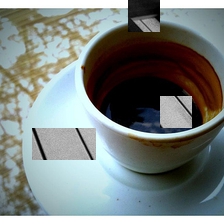}
    \end{subfigure}
    \begin{subfigure}{.15\textwidth}
        \centering
        \includegraphics[width=\linewidth, frame]{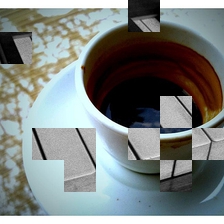}
    \end{subfigure}
    \begin{subfigure}{.15\textwidth}
        \centering
        \includegraphics[width=\linewidth, frame]{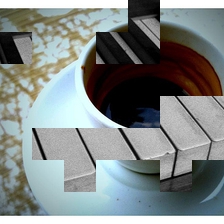}
    \end{subfigure}
    \begin{subfigure}{.15\textwidth}
        \centering
        \includegraphics[width=\linewidth, frame]{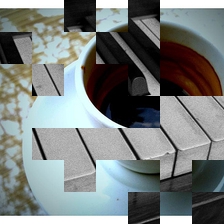}
    \end{subfigure}
    \begin{subfigure}{.15\textwidth}
        \centering
        \includegraphics[width=\linewidth, frame]{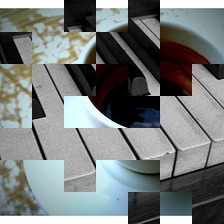}
    \end{subfigure}
    \caption{Examples of patch mixing experiments with (14, 14) - top - and (7, 7) - bottom - grid sizes with increasing information loss from 10-50\% in decadal intervals.}
    \label{fig:patch_mixing}
\end{figure}

Figure~\ref{fig:patch_mixing} illustrates the patch mixing experiments for both (7, 7) and (14, 14) grid sizes as the number of out-of-context patches increases. This is discussed in Section 4.1 of the main paper. ConvNeXt trained with Patch Mixing performs better than the original model for all experiments.

\section*{Random Patch Drop}
\begin{figure}
    \centering
    \centering
    \resizebox{0.7\textwidth}{!}{%
    \includegraphics[width=0.3\linewidth]{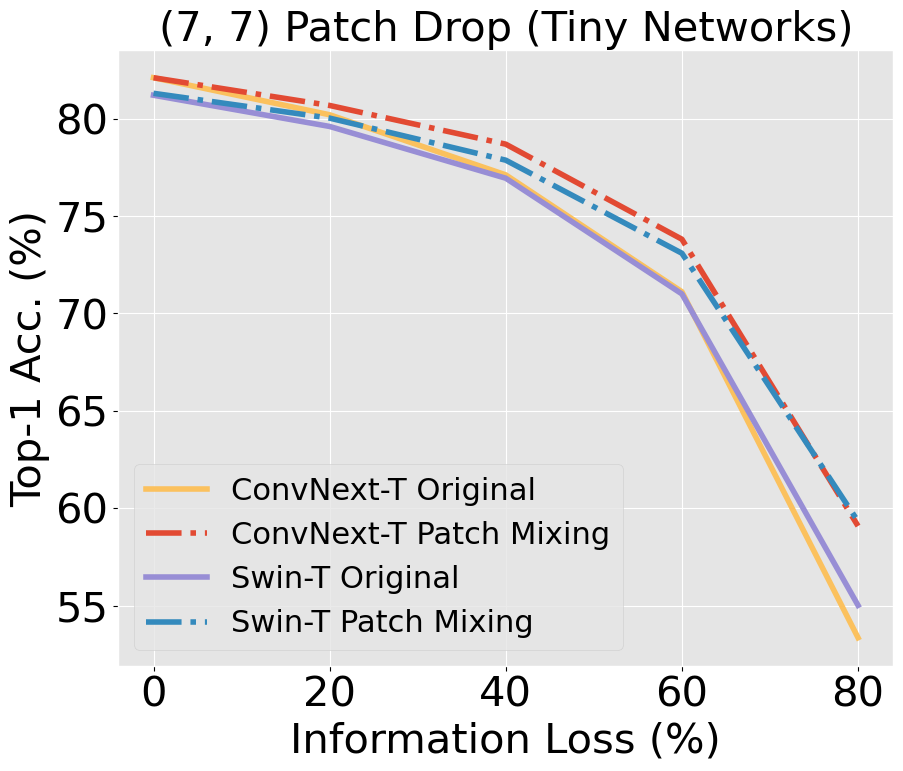}
    \includegraphics[width=0.3\linewidth]{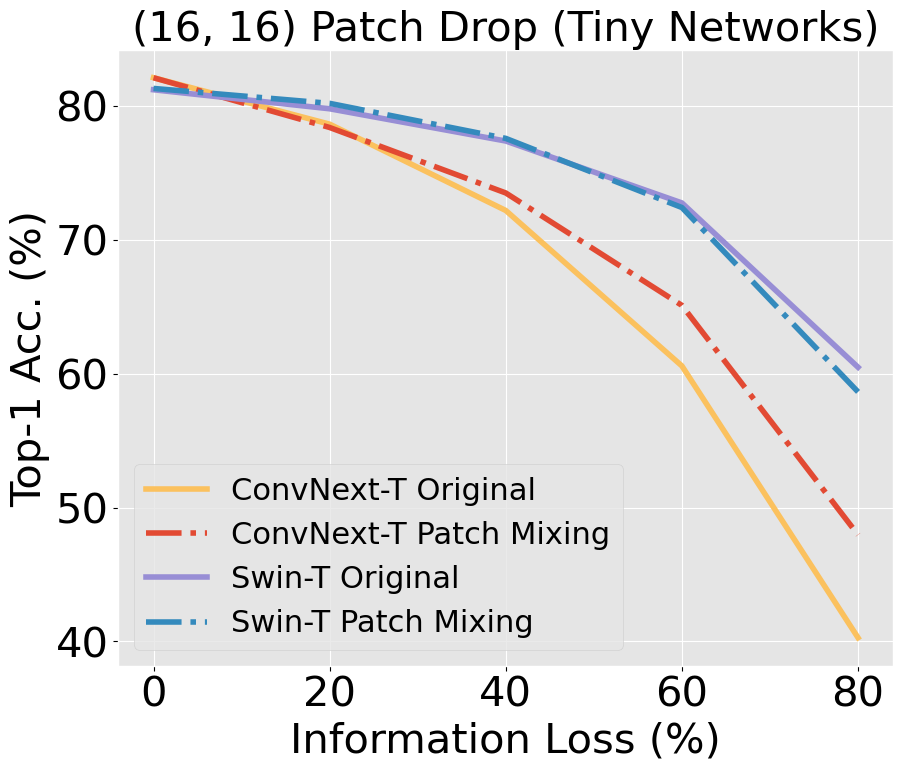}
    }
    \resizebox{0.7\textwidth}{!}{%
    \includegraphics[width=0.3\linewidth]{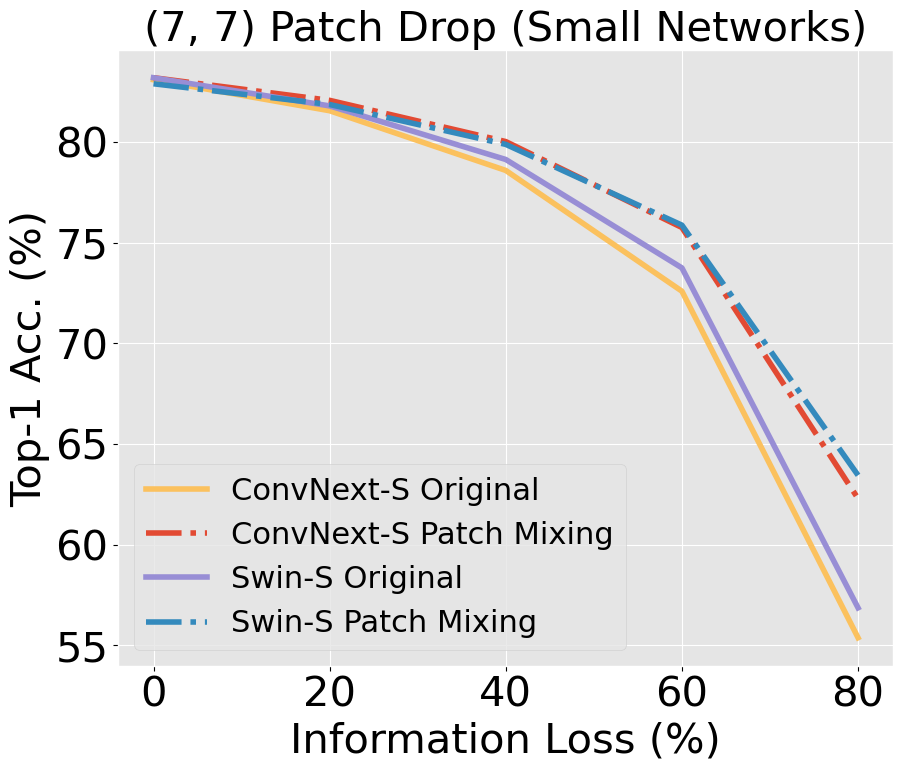}
    \includegraphics[width=0.3\linewidth]{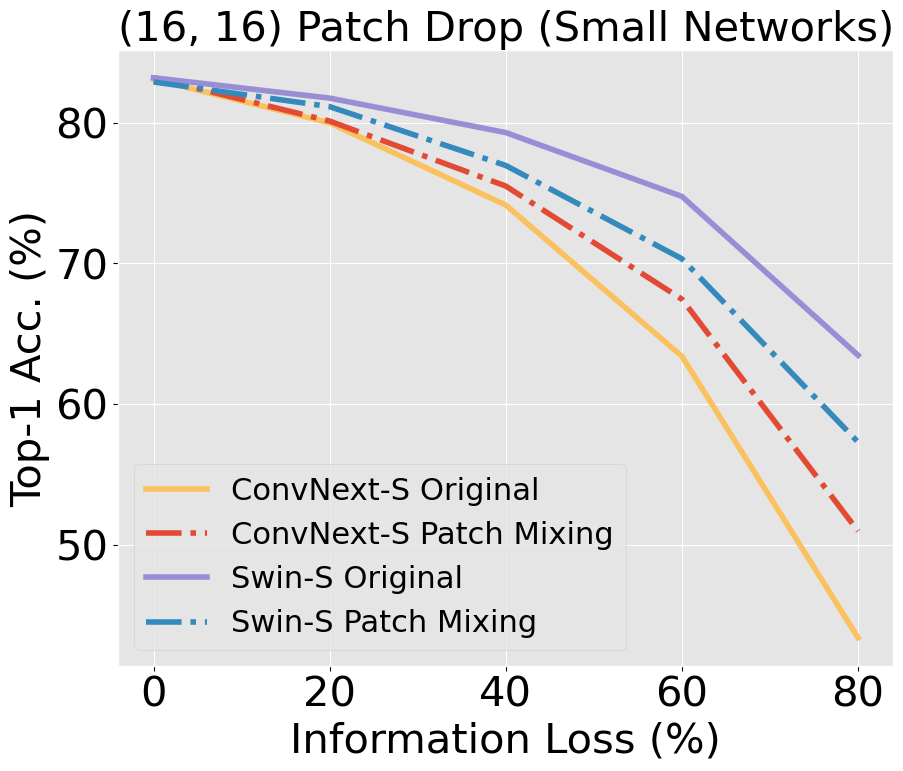}
    }
    \caption{Random patch drop: tiny and small networks}
    \label{fig:small_drop2}
\end{figure}

In this section we provide additional patch drop experiments for (7, 7) and (16, 16) grid sizes, illustrated in Figure~\ref{fig:small_drop2}. Similar to the results of the (14, 14) grid shown in Section 4.3 of the main paper, ConvNeXt trained with Patch Mixing outperforms its counterpart in all cases. Additionally, for the (7, 7) grid we see that ConvNeXt Patch Mixing outperforms the original Swin Models for Tiny and Small networks and is on par with Swin Patch Mixing. We also see that Swin performance either stays static or slightly increases, but not by the same magnitude as ConvNext performance.

\begin{figure}
    \centering
    \begin{tabular}{cccc}
        \multicolumn{4}{c}{Information loss} \\
        20\% & 40\% & 60\% & 80\% \\
        \begin{subfigure}{.18\textwidth}
            \centering
            \includegraphics[width=\linewidth, frame]{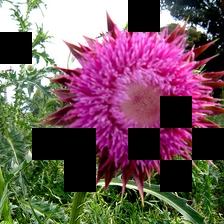}
        \end{subfigure} & 
        \begin{subfigure}{.18\textwidth}
            \centering
            \includegraphics[width=\linewidth, frame]{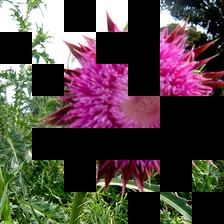}
        \end{subfigure} &
        \begin{subfigure}{.18\textwidth}
            \centering
            \includegraphics[width=\linewidth, frame]{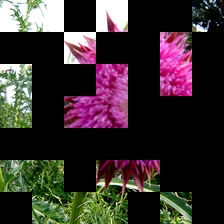}
        \end{subfigure} &
        \begin{subfigure}{.18\textwidth}
            \centering
            \includegraphics[width=\linewidth, frame]{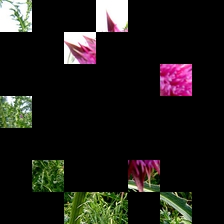}
        \end{subfigure} \\
        \begin{subfigure}{.18\textwidth}
            \centering
            \includegraphics[width=\linewidth, frame]{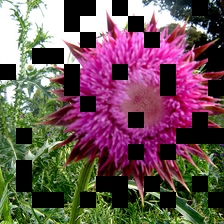}
        \end{subfigure} & 
        \begin{subfigure}{.18\textwidth}
            \centering
            \includegraphics[width=\linewidth, frame]{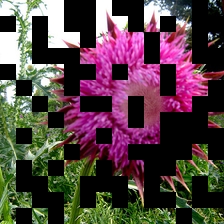}
        \end{subfigure} &
        \begin{subfigure}{.18\textwidth}
            \centering
            \includegraphics[width=\linewidth, frame]{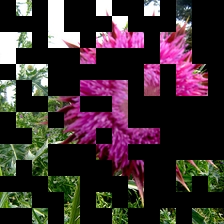}
        \end{subfigure} &
        \begin{subfigure}{.18\textwidth}
            \centering
            \includegraphics[width=\linewidth, frame]{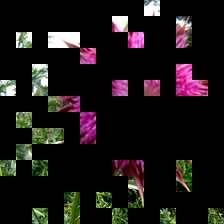}
        \end{subfigure} \\
        \begin{subfigure}{.18\textwidth}
            \centering
            \includegraphics[width=\linewidth, frame]{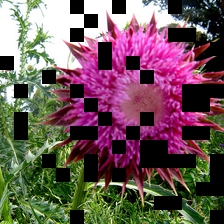}
        \end{subfigure} & 
        \begin{subfigure}{.18\textwidth}
            \centering
            \includegraphics[width=\linewidth, frame]{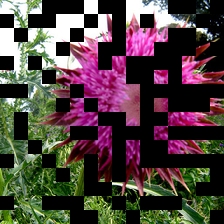}
        \end{subfigure} &   
        \begin{subfigure}{.18\textwidth}
            \centering
            \includegraphics[width=\linewidth, frame]{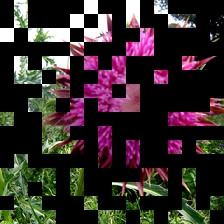}
        \end{subfigure} &
        \begin{subfigure}{.18\textwidth}
            \centering
            \includegraphics[width=\linewidth, frame]{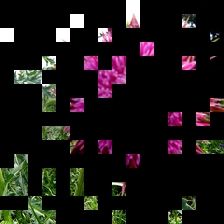}
        \end{subfigure}
    \end{tabular}
    \caption{Patch drop examples for the (7,7), (14,14), and (16,16) grid sizes in the top, middle, and bottom rows, respectively.}
    \label{fig:patchdrop}
\end{figure}

Figure~\ref{fig:patchdrop} provides a visualization of the random patch drop~\cite{intriguing_transformers} experiments conducted for (7,7), (14,14), and (16,16) grids with increasing information loss up to 80\%. 

\section*{Patch permutations}

\begin{figure}[H]
    \centering
    \begin{subfigure}{.18\textwidth}
        \centering
        \includegraphics[width=\linewidth, frame]{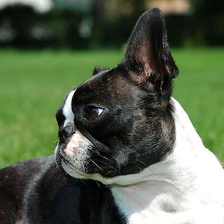}
    \end{subfigure}
    \hfill 
    \begin{subfigure}{.18\textwidth}
        \centering
        \includegraphics[width=\linewidth, frame]{perm_mixed_2.jpg}
    \end{subfigure}
    \begin{subfigure}{.18\textwidth}
        \centering
        \includegraphics[width=\linewidth, frame]{perm_mixed_4.jpg}
    \end{subfigure}
    \begin{subfigure}{.18\textwidth}
        \centering
        \includegraphics[width=\linewidth, frame]{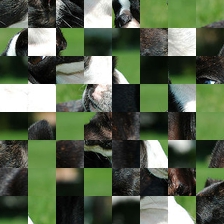}
    \end{subfigure}
    \begin{subfigure}{.18\textwidth}
        \centering
        \includegraphics[width=\linewidth, frame]{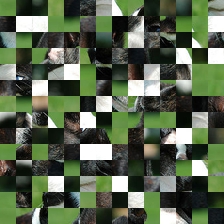}
    \end{subfigure}
    \caption{Examples of patch permutations. The sequence from left to right is: the original, unaltered image, followed by images with shuffled patches. The total shuffle grid sizes for these subsequent images are 4, 16, 64, and 196, respectively.}
    \label{fig:perms2}
\end{figure}

Figure~\ref{fig:perms2} illustrates the patch permutation experiments~\cite{intriguing_transformers} discussed in Section 4.2 of the main paper. Images are shown with increasing shuffle grid size, which is the total quantity of image patches in the shuffled images. The performance gap between original and Patch Mixing trained ConvNeXt models widens with increasing shuffle grid size, reaching over 20\%, while the gap between ConvNeXt-T trained with Patch Mixing and the original Swin-T remains negligible, even with larger shuffle grid sizes. 

\section*{c-RISE}

\begin{figure}[H]
    \centering
    \includegraphics[width=\linewidth]{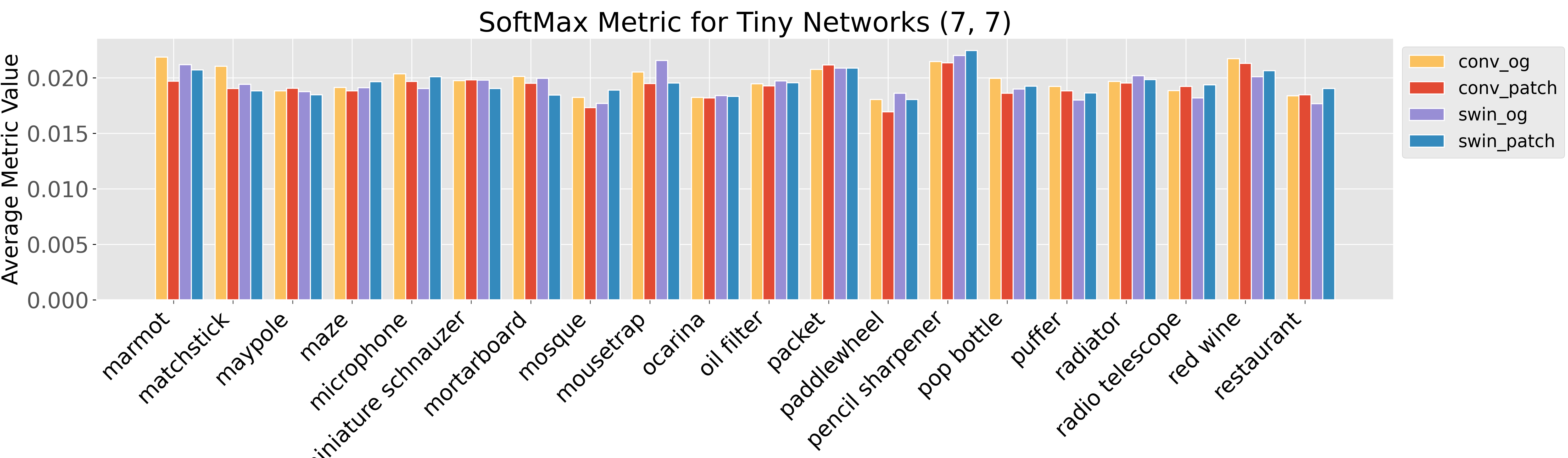}
    \caption{Average out-of-context SoftMax metric by class.}
    \label{fig:c-rise}
\end{figure}

In Figure~\ref{fig:c-rise} we illustrate the average out-of-context Softmax metric by class for a random subset of 20 classes from the 100 classes tested. The methods and results of c-RISE are discussed in Sections 2.2 and 4.3 of the paper, respectively. ConvNeXt Patch Mixing performs better than the original model for all but 3 classes.

\section*{Broader Impact Discussion}

This research paper shares both the benefits and potential risks inherent in advancing the discriminative behavior of computer vision models. Our refinement of the Patch Mixing technique improves robustness of CNNs to occlusion, making them more useful in real-world applications where partial visibility of objects is common, such has autonomous vehicles, security cameras, and biomedical imaging. It also extends our understanding of how ViTs and CNNs differ in their treatment of out-of-context information, which could have implications for their application in settings requiring robustness to occlusion. Unfortunately, this method can also be used to compromise models by exploiting the inductive bias linked to patch selectivity.

A key aspect of this research is the development of the c-RISE explainability technique. c-RISE provides a valuable tool for machine learning researchers seeking to distinguish evidence between the top-1 prediction and the remaining classes. Such insights can help in debugging, fine-tuning, and generally improving the reliability and fairness of AI systems, particularly in sensitive areas like healthcare, where interpretability of AI decisions are incredibly important.

\end{document}